\pdfoutput=1

\PassOptionsToPackage{svgnames, table}{xcolor}

\documentclass[11pt]{article}

\usepackage{acl}

\usepackage{times}
\usepackage{latexsym}

\usepackage[T1]{fontenc}

\usepackage[utf8]{inputenc}

\usepackage{microtype}

\usepackage{inconsolata}

\usepackage{graphicx}

\usepackage{amsmath}
\usepackage{amssymb}
\usepackage{arydshln}
\usepackage{booktabs}
\usepackage{enumitem}
\usepackage{multirow}
\usepackage{pifont}
\usepackage{soul}
\usepackage{xurl}

\newcommand{\cmark}{\textcolor{ForestGreen}{\ding{51}}}
\newcommand{\xmark}{\textcolor{FireBrick}{\ding{55}}}

\newcommand{\visiblenewline}{\textcolor{gray!50}{\textbackslash n}}
\newcommand{\specialtoken}[1]{\textcolor{gray!50}{\texttt{#1}}}

\colorlet{best}{MediumAquamarine!40}
\newcommand{\highlight}[1]{\sethlcolor{best}\hl{#1}}

\newcommand{\result}[2]{#1\textsubscript{\,\textcolor{gray}{#2}}}
\newcommand{\bestresult}[2]{\cellcolor{best} \result{\textbf{#1}}{#2}}
\newcommand{\bestcatresult}[2]{\cellcolor{best} \result{#1}{#2}}

\makeatletter
  \def\adl@drawiv#1#2#3{%
    \hskip.5\tabcolsep
    \xleaders#3{#2.5\@tempdimb #1{1}#2.5\@tempdimb}%
    #2\z@ plus1fil minus1fil\relax
    \hskip.5\tabcolsep
  }
  \newcommand{\cdashlinelr}[1]{%
    \noalign{\vskip\aboverulesep
      \global\let\@dashdrawstore\adl@draw
      \global\let\adl@draw\adl@drawiv}
    \cdashline{#1}[0.5pt/2pt]
    \noalign{\global\let\adl@draw\@dashdrawstore
      \vskip\belowrulesep}
  }
\makeatother

\title{Self-calibration for Language Model Quantization and Pruning}

\author{Miles Williams$^{\diamondsuit \spadesuit}$ \quad George Chrysostomou$^{\spadesuit}$ \quad Nikolaos Aletras$^{\diamondsuit}$ \\ $^{\diamondsuit}$University of Sheffield\\$^{\spadesuit}$Enterprise AI Services, AstraZeneca\\ \texttt{\{mwilliams15, n.aletras\}@sheffield.ac.uk}}

\begin{document}
\maketitle
\begin{abstract}
Quantization and pruning are fundamental approaches for model compression, enabling efficient inference for language models. In a post-training setting, state-of-the-art quantization and pruning methods require calibration data, a small set of unlabeled examples. Conventionally, this is randomly sampled web text, aiming to reflect the model training data. However, this poses two key problems: (1) unrepresentative calibration examples can harm model performance, and (2) organizations increasingly avoid releasing model training data. In this paper, we propose self-calibration as a solution. Our approach requires no external data, instead leveraging the model itself to generate synthetic calibration data, with a view to better approximating the pre-training data distribution. We extensively compare the performance of self-calibration with several baselines, across a variety of models, compression methods, and tasks. Our approach proves consistently competitive in maximizing downstream task performance, frequently outperforming even using real data.\footnote{\url{https://github.com/mlsw/llm-compression-calibration}}
\end{abstract}

\section{Introduction}

Large language models (LLMs) trained using vast corpora have delivered remarkable advances across a variety of domains and tasks \citep{touvron-etal-2023-llama, jiang-etal-2023-mistral, gemma-team-etal-2024-gemma}. However, they demand extensive computational resources for inference \citep{wu-etal-2022-sustainable, luccioni-etal-2023-estimating}, presenting a limiting factor for their practical use. Consequently, this has prompted the development of an extensive collection of methods to improve inference efficiency \citep{treviso-etal-2023-efficient}. In particular, model compression aims to reduce the size of a model while retaining downstream task performance \citep{wan-etal-2024-efficient}.

Quantization and pruning have emerged as prominent model compression approaches for LLMs \citep{gholami-etal-2021-survey, wan-etal-2024-efficient}. Pruning removes less important weights from the model, while quantization represents the weights (and possibly activations) using fewer bits. Both quantization and pruning can be effectively applied in a post-training setting, retaining comparable performance across a range of downstream tasks \citep{frantar-etal-2023-optq, frantar-alistarh-2023-sparsegpt, sun-etal-2024-simple, lin-etal-2024-awq}.

\begin{figure}
\centering
\includegraphics[scale=0.5]{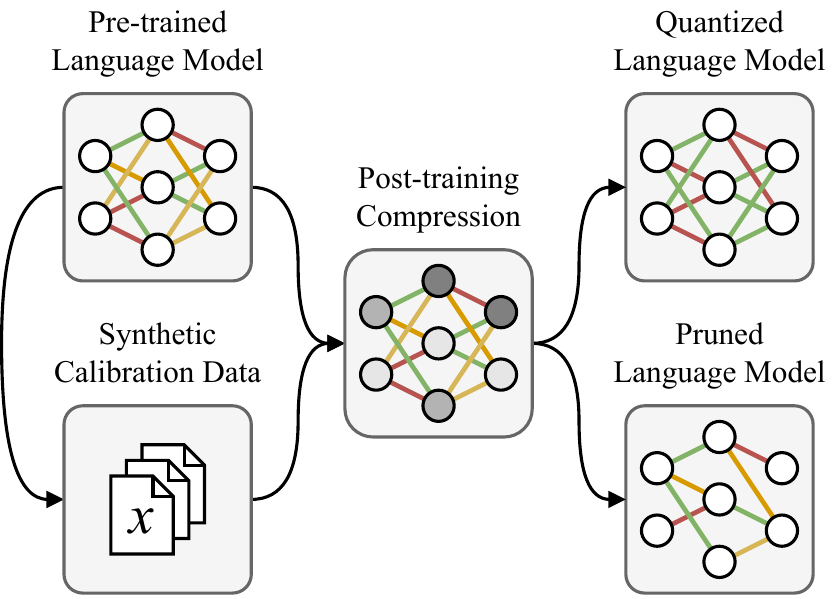}
\caption{Self-calibration for the post-training quantization and pruning of language models.}
\label{fig:enter-label}
\end{figure}

Post-training quantization and pruning typically depend upon \textit{calibration data}, a small set of unlabeled examples \citep{nagel-etal-2020-up, hubara-etal-2021-accurate} used to generate layer activations throughout the model. Conventionally, LLM calibration data consists of randomly sampled web text \citep{frantar-etal-2023-optq, sun-etal-2024-simple, lin-etal-2024-awq}, aiming to reflect the model training data distribution.

However, recent work has questioned the influence of calibration data in LLM compression. \citet{jaiswal-etal-2024-compressing} hint that the careful selection of calibration data may benefit high-sparsity pruning. Concurrently, \citet{williams-aletras-2024-impact} illustrate the impact of calibration data in quantization and pruning. Finally, \citet{zeng-etal-2024-multilingual} and \citet{kurz-etal-2024-investigating} highlight the role of language-specific calibration data for multilingual models.

To further complicate matters, organizations are increasingly reluctant to release model training data or disclose necessary replication details. Table \ref{tab:large_language_models} illustrates that although the weights of some state-of-the-art LLMs are openly available, their training data is largely unavailable. This may be due to (1) legal liability concerns arising from data licensing \citep{eckart-de-castilho-etal-2018-legal}, and (2) privacy concerns when using  proprietary or personal data \citep{carlini-etal-2021-extracting}. Moreover, publicly released training data can later become unavailable. For example, The Pile \citep{gao-etal-2020-pile} is no longer distributed due to copyright violations. The absence of training data raises the question of how representative calibration data can be selected, when the training distribution itself is unknown. This issue is especially relevant for models trained primarily with private datasets, such as Microsoft's Phi series of models \citep{gunasekar-etal-2023-textbooks, li-etal-2023-textbooks}.

In this paper, we propose self-calibration as a solution to concerns surrounding the availability and quality of calibration data. Our approach removes the need for external calibration data sources, instead leveraging the model itself to automatically generate synthetic calibration data. We compare our approach to various real and synthetic datasets, including data sampled from a large mixture-of-experts model. Our approach is consistently competitive in maximizing the performance of compressed models, across a variety of models and compression methods. In many cases, we find that self-calibration can outperform even real data.

\section{Related Work}

\subsection{Model Compression}

Model compression aims to reduce the size of a model without compromising downstream task performance, therefore reducing the computational resources required for inference \citep{treviso-etal-2023-efficient}. Quantization and pruning are two prominent model compression approaches that have been widely applied to LLMs \citep{wan-etal-2024-efficient}.

\paragraph{Pruning.}

The goal of pruning is to remove redundant model weights \citep{lecun-etal-1989-optimal}. Pruning often relies upon a fine-tuning step \citep{han-etal-2015-learning, sanh-etal-2020-movement}, however this is challenging at the scale of LLMs. Alternatively, there have been various efforts towards adapting the Optimal Brain Surgeon (OBS) framework \citep{lecun-etal-1989-optimal, hassibi-etal-1993-optimal} for language model pruning \citep{frantar-etal-2021-m-fac, kurtic-etal-2022-optimal, frantar-alistarh-2022-optimal}. However, the extensive size of LLMs makes it impractical to apply such methods. SparseGPT \citep{frantar-alistarh-2023-sparsegpt} presents an approximate weight reconstruction approach, enabling efficient LLM pruning without compromising performance. Separately, Wanda \citep{sun-etal-2024-simple} relies on a pruning criterion that does not require second-order information, allowing pruning with a single forward pass.

\paragraph{Quantization.}

The aim of quantization is to represent model weights (and potentially activations) using fewer bits. Large-magnitude outlier features pose a significant problem for the quantization of LLMs, which can be addressed through holding these in higher precision \citep{dettmers-etal-2022-gpt3}. However, this approach is less hardware-friendly. Instead, SmoothQuant \citep{xiao-etal-2023-smoothquant} migrates the difficulty of activation quantization to the weights, which are easier to quantize. AWQ \citep{lin-etal-2024-awq} presents a hardware-friendly approach for holding a small fraction of the weights in higher precision. In a separate line of work, \citet{frantar-alistarh-2022-optimal} adapt the OBS framework to quantization. GPTQ \citep{frantar-etal-2023-optq} builds upon this work to enable second-order low-bit quantization for LLMs.

\begin{table}[t]
\small
\centering
\begin{tabular}{llcc}
\toprule
& & \multicolumn{2}{c}{Open Source} \\
\cmidrule(lr){3-4}
Model & Reference &  Weights & Data \\
\midrule
GPT-4 & \citet{openai-etal-2024-gpt4} & \xmark & \xmark \\
Mistral & \citet{jiang-etal-2023-mistral} & \cmark & \xmark \\
Llama 2 & \citet{touvron-etal-2023-llama2} & \cmark & \xmark \\
Falcon & \citet{almazrouei-etal-2023-falcon} & \cmark & \cmark \\
Phi-2 & \citet{javaheripi-etal-2023-phi} & \cmark & \xmark \\
Gemini & \citet{gemini-team-etal-2024-gemini} & \xmark & \xmark \\
OLMo & \citet{groeneveld-etal-2024-olmo} & \cmark & \cmark \\
Claude 3 & \citet{anthropic-2024-claude} & \xmark & \xmark \\
Gemma & \citet{gemma-team-etal-2024-gemma} & \cmark & \xmark \\
\bottomrule
\end{tabular}
\caption{The training data for state-of-the-art LLMs is rarely available. Models selected according to benchmark performance and ordered by publication date.}
\label{tab:large_language_models}
\end{table}

\subsection{Calibration Data}
\label{sec:calibration-data}

In a post-training setting, model compression methods rely upon calibration data \citep{wan-etal-2024-efficient}. This consists of a small set of unlabeled examples, used to generate layer activations \citep{nagel-etal-2020-up, hubara-etal-2021-accurate}. Calibration data for LLMs conventionally consists of text sampled from a curated training dataset \citep{frantar-etal-2023-optq, xiao-etal-2023-smoothquant, frantar-alistarh-2023-sparsegpt, sun-etal-2024-simple, lin-etal-2024-awq}. In practice, the exact model training data may not be publicly available (Table \ref{tab:large_language_models}). Consequently, large scale web text datasets (e.g. C4; \citealp{raffel-etal-2020-exploring}) are ordinarily used as an approximation of the pre-training distribution. Recent work has questioned the performance impact of the calibration data used for LLM compression \citep{jaiswal-etal-2024-compressing, williams-aletras-2024-impact, zeng-etal-2024-multilingual}. Synthetic data presents a promising avenue towards alleviating such concerns, including the varied quality of web text examples \citep{dodge-etal-2021-documenting}. However, synthetic calibration data for post-training LLM compression has yet to be systematically explored.

Synthetic data for model compression has been previously explored in computer vision, regularly motivated by privacy and security concerns arising from sensitive training images (e.g. medical contexts). \citet{haroush-etal-2020-knowledge} and \citet{cai-etal-2020-zeroq} proposed approaches for data-free quantization \citep{nagel-etal-2019-data}, allowing the model itself to synthesize input data for quantization. Fundamentally, these approaches generate images matching the learned statistics from batch normalization layers \citep{zhang-etal-2021-diversifying, li-etal-2023-hard}, which are notably absent in LLMs \citep{wang-etal-2022-understanding-failure}.

\subsection{Synthetic Data with Language Models}

Synthetic data refers to artificial data that has been created with the aim of imitating real-world data \citep{liu-etal-2024-best}. In the context of language models, supervised training of classification models with synthetic labeled data has been widely explored \citep{kumar-etal-2020-data, schick-schutze-2021-generating, sahu-etal-2022-data, meng-etal-2022-generating, chung-etal-2023-increasing, li-etal-2023-synthetic}. Similarly, synthetic data has seen broad use for supervised instruction fine-tuning \citep{wang-etal-2023-self-instruct, ding-etal-2023-enhancing, xu-etal-2024-wizardlm}. Most recently, partially or entirely synthetic datasets have been used for pre-training \citep{gunasekar-etal-2023-textbooks, li-etal-2023-textbooks, maini-etal-2024-rephrasing, benallal-etal-2024-cosmopedia}. However, the distribution of such datasets may deviate from the pre-training distribution of other LLMs.

\section{Self-calibration}

When the exact training data for a model is unavailable, sampling calibration data from an alternative distribution offers an approximation at best. Even if the exact training data is available, individual examples may be noisy and deviate from the overall distribution. To address these limitations, we propose self-calibration, a general-purpose adaptation to model compression that relies on calibration data from the model itself. Our hypothesis is that sampling from the learned posterior distribution, which approximates the training data, offers more representative calibration examples. In turn, we expect that such calibration examples will enable greater preservation of downstream task performance following model compression.

\subsection{Synthesizing Calibration Data}
\label{sec:synthesizing-calibration-data}

We formulate the synthesis of calibration examples as an open-ended text generation problem for a specific language model that we wish to compress. Crucially, we aim to generate synthetic data that is as representative as possible with respect to the training distribution. To achieve this, we refrain from using external data, which introduces assumptions about the training data distribution.

Fundamentally, text generation consists of predicting the next token in a sequence. Formally, we compute a probability distribution over the vocabulary $\mathcal{V}$ for the next token $w_i$, given context $w_{1:i-1}$. Taking the context as input, a language model generates the output logits, $u_{1:|\mathcal{V}|}$. The probability distribution is then formed through normalizing the logits with the softmax function.

To generate calibration data that reflects the model training data distribution, we condition generation upon only the beginning-of-sequence token (e.g. \texttt{<s>} or \texttt{<|start\_of\_text|>}). We continue to generate tokens until either the end-of-sequence token or maximum sequence length is reached. In the event that a generation does not reach the desired length, we simply concatenate additional generations. As a prefix or prompt would introduce bias and require external data, we do not directly condition generation. Instead, we rely upon scheduled temperature sampling to guide generation.

\subsection{Temperature Scheduling}

The softmax function can be additionally parameterized with a temperature $t$, to control the sharpness of the probability distribution \citep{ackley-etal-1985-learning, hinton-etal-2015-distilling}. A lower temperature concentrates the probability mass on the most likely tokens, while a higher temperature disperses the probability mass more uniformly. In practice, the temperature influences characteristics of the generated text, often improving its quality and diversity compared to greedy decoding \citep{holtzman-etal-2020-curious, meister-etal-2023-locally}.

When generating text without context, we hypothesize that the first few generated tokens are crucial, influencing the content and coherence. To explore a variety of prefixes, we propose the use of a temperature schedule, inspired by \citet{carlini-etal-2021-extracting}. Formally, we define the probability of a token as: $$P(w_i \mid w_{1:i-1}) = \frac{\exp(u_i/t_i)}{\sum_{j=1}^{|\mathcal{V}|} \exp(u_j/t_i)}$$

\noindent where $t_i$ scales linearly from $t_\text{initial}$ at the start of generation to $t_\text{final}$, across $n$ token generation steps: $$t_i = \begin{cases}
t_\text{initial} + \dfrac{i}{n}(t_\text{final} - t_\text{initial}) & \text{if } i \leq n, \\
t_\text{final} & \text{if } i > n.
\end{cases}$$

In practice, a temperature schedule enables us to experiment with a variety of generation strategies. For example, we are able to generate a diverse prefix (i.e. $t_\text{initial} > 1$) followed by a more confident continuation (i.e. $t_\text{final} \leq 1$), as well as a high-likelihood prefix followed by a creative continuation. We provide a comprehensive ablation of these parameters choices in \S \ref{sec:sampling-strategy-ablation}. For comparison, we also present results with greedy decoding and standard sampling (i.e. without temperature).

\section{Experimental Setup}

\subsection{Baseline Calibration Data}
\label{sec:calibration-data-sources}

\paragraph{Real data.}

To evaluate the performance of self-calibration for LLM compression, we first consider real-world datasets that are conventionally used for LLM compression \citep{frantar-etal-2023-optq}.

\begin{itemize}[left=0pt]
\item \textbf{C4} \citep{raffel-etal-2020-exploring}: The Colossal Clean Crawled Corpus is routinely used as a source of calibration data (\S \ref{sec:calibration-data}). This consists of web-text that has been deduplicated and filtered to maximize high-quality natural language text.
\item \textbf{WikiText} \citep{merity-etal-2017-pointer}: The WikiText dataset consists of a high-quality encyclopedic text from Wikipedia. Notably, this includes only articles highlighted as `Good' or `Featured' by human editors. The review process assesses accuracy and writing quality, amongst other factors.
\end{itemize}

\paragraph{Synthetic data.}

Separately, we compare the performance of self-calibration with synthetic data generated (1) without a language model, and (2) with a substantially larger external model.

\begin{itemize}[left=0pt]
\item \textbf{Vocabulary}: As a simple baseline, we create examples consisting of tokens randomly sampled from the model vocabulary. We assume a uniform distribution over the vocabulary, however we exclude special purpose tokens (e.g. \texttt{<unk>}).
\item \textbf{Cosmopedia} \citep{benallal-etal-2024-cosmopedia}: The Cosmopedia dataset consists of a broad range of synthetic text, including textbooks, blog posts, and stories. These were created by prompting Mixtral 8x7B Instruct \citep{jiang-etal-2024-mixtral} with a variety of high-quality topics selected from real data.
\end{itemize}

\begin{table*}[t]
\scriptsize
\centering
\begin{tabular}{lllrrrrr}
\toprule
Method & Type & Calibration Dataset & \multicolumn{1}{c}{Gemma 2B} & \multicolumn{1}{c}{Phi-2 2.7B} & \multicolumn{1}{c}{OPT 6.7B} & \multicolumn{1}{c}{Mistral 7B} & \multicolumn{1}{c}{Llama 3.1 8B} \\
\midrule
- & \multicolumn{2}{l}{-} & \result{60.7}{\hphantom{0.0}} & \result{65.8}{\hphantom{0.0}} &  \result{57.4}{\hphantom{0.0}} & \result{67.4}{\hphantom{0.0}} & \result{67.8}{\hphantom{0.0}} \\
\cmidrule{1-8}
\multirow[c]{5}{*}{AWQ} & \multirow[c]{2}{*}{Real} & C4 & \result{59.5}{0.2} & \result{65.4}{0.2} & \result{57.6}{0.1} & \result{67.1}{0.0} & \result{66.9}{0.2} \\
 &  & WikiText & \result{59.5}{0.2} & \result{65.4}{0.2} & \result{57.5}{0.1} & \result{67.1}{0.1} & \result{67.1}{0.1} \\
\cdashlinelr{2-8}
 & \multirow[c]{3}{*}{Synthetic} & Vocabulary & \result{59.3}{0.2} & \result{64.5}{0.2} & \result{56.6}{0.3} & \result{66.5}{0.1} & \result{66.0}{0.3} \\
 &  & Cosmopedia & \result{59.8}{0.2} & \result{65.3}{0.2} & \result{57.6}{0.1} & \result{67.0}{0.2} & \result{66.9}{0.3} \\
 &  & Self-calibration (Ours) & \bestresult{59.8}{0.4} & \bestresult{65.4}{0.2} & \bestresult{57.6}{0.1} & \bestcatresult{67.0}{0.2} & \result{66.6}{0.3} \\
\cmidrule{1-8}
\multirow[c]{5}{*}{GPTQ} & \multirow[c]{2}{*}{Real} & C4 & \result{58.7}{0.4} & \result{64.7}{0.3} & \result{56.8}{0.2} & \result{66.8}{0.3} & \result{66.9}{0.3} \\
 &  & WikiText & \result{58.6}{0.3} & \result{64.6}{0.2} & \result{56.9}{0.1} & \result{66.9}{0.3} & \result{66.6}{0.3} \\
\cdashlinelr{2-8}
 & \multirow[c]{3}{*}{Synthetic} & Vocabulary & \result{57.9}{0.3} & \result{64.3}{0.2} & \result{56.6}{0.3} & \result{66.0}{0.1} & \result{65.7}{0.1} \\
 &  & Cosmopedia & \result{58.5}{0.3} & \result{64.3}{0.1} & \result{56.8}{0.1} & \result{66.6}{0.2} & \result{66.9}{0.1} \\
 &  & Self-calibration (Ours) & \bestresult{59.9}{0.3} & \bestresult{65.0}{0.3} & \bestresult{56.9}{0.2} & \result{65.9}{0.2} & \result{66.1}{0.3} \\
\cmidrule{1-8}
\multirow[c]{5}{*}{SparseGPT} & \multirow[c]{2}{*}{Real} & C4 & \result{49.7}{0.8} & \result{54.3}{0.3} & \result{52.8}{0.2} & \result{57.3}{0.3} & \result{54.8}{0.3} \\
 &  & WikiText & \result{48.3}{0.2} & \result{53.3}{0.5} & \result{51.6}{0.2} & \result{55.5}{0.3} & \result{52.6}{0.4} \\
\cdashlinelr{2-8}
 & \multirow[c]{3}{*}{Synthetic} & Vocabulary & \result{43.4}{0.3} & \result{50.1}{0.2} & \result{47.7}{0.2} & \result{53.0}{0.4} & \result{47.3}{0.4} \\
 &  & Cosmopedia & \result{47.7}{0.3} & \result{52.3}{0.2} & \result{50.9}{0.2} & \result{55.1}{0.3} & \result{50.9}{0.3} \\
 &  & Self-calibration (Ours) & \bestresult{50.8}{0.2} & \bestresult{56.4}{0.3} & \bestcatresult{52.7}{0.3} & \bestcatresult{56.8}{0.3} & \bestcatresult{53.8}{0.4} \\
\cmidrule{1-8}
\multirow[c]{5}{*}{Wanda} & \multirow[c]{2}{*}{Real} & C4 & \result{44.2}{0.2} & \result{50.4}{0.4} & \result{50.6}{0.2} & \result{53.7}{0.3} & \result{49.0}{0.3} \\
 &  & WikiText & \result{44.8}{0.4} & \result{49.9}{0.2} & \result{49.2}{0.2} & \result{53.4}{0.2} & \result{49.2}{0.1} \\
\cdashlinelr{2-8}
 & \multirow[c]{3}{*}{Synthetic} & Vocabulary & \result{42.1}{0.4} & \result{47.0}{0.3} & \result{43.2}{0.1} & \result{48.4}{0.2} & \result{44.7}{0.3} \\
 &  & Cosmopedia & \result{44.5}{0.2} & \result{49.4}{0.4} & \result{48.7}{0.2} & \result{52.7}{0.2} & \result{47.7}{0.2} \\
 &  & Self-calibration (Ours) & \bestresult{45.2}{0.3} & \bestresult{51.5}{0.7} & \bestresult{50.7}{0.2} & \bestcatresult{53.5}{0.1} & \bestcatresult{49.1}{0.1} \\
\bottomrule
\end{tabular}
\caption{Average task accuracy across five calibration sets for all models, with standard deviation denoted in subscript. \highlight{Highlighted} values indicate that self-calibration (ours) matches or exceeds the performance of all synthetic datasets. \textbf{Bold} values additionally indicate that self-calibration matches or exceeds the highest performing dataset overall, including the real datasets.
Self-calibration temperature is fixed at 1.0 to enable fair comparison.}
\label{tab:task_results}
\end{table*}

\paragraph{Sampling.}

Following convention, we randomly sample 128 calibration examples consisting of 2,048 tokens each \citep{frantar-etal-2023-optq, frantar-alistarh-2023-sparsegpt, sun-etal-2024-simple, chrysostomou-etal-2024-investigating}. Although the aim of random sampling is to avoid selection bias, it could produce a sample that is less representative of the source dataset. Consequently, we repeat the sampling process to create five distinct calibration sets for each source dataset. We present an ablation study on the quantity of calibration data used in \S \ref{sec:data-quantity-ablation}.

Certain models (Gemma, Mistral, and Llama) were trained using multilingual data, which is reflected when sampling from these models. To enable a fair comparison with our English-only calibration datasets and evaluation tasks, we promote the generation of English-language text for these models. Specifically, we constrain only the first generation step to a pre-defined list of English stop words curated by \citet{honnibal-etal-2020-spacy}.

\subsection{Models}
\label{sec:models}

We experiment with popular `open-source' LLMs from five different model families: (1) \textbf{Gemma 2B} \citep{gemma-team-etal-2024-gemma}, (2) \textbf{Phi-2 2.7B} \citep{javaheripi-etal-2023-phi}, (3) \textbf{OPT 6.7B} \citep{zhang-etal-2022-opt}, (4) \textbf{Mistral 7B} (v0.3) \citep{jiang-etal-2023-mistral}, and (5) \textbf{Llama 3.1 8B} \citep{dubey-etal-2024-llama3}.\footnote{
\citet{gemma-team-etal-2024-gemma} use a naming scheme that excludes embedding parameters. For comparison, we note that Gemma 2B has 2.5B trainable parameters. We also note that the embedding parameters are shared \citep{press-wolf-2017-using}.} 

With the exception of OPT, which was pre-trained using only publicly available datasets, limited details surrounding the training data distribution have been disclosed. The training data for all models is reported to include public web documents. However, the training data for Phi-2 notably relies upon a substantial proportion of synthetic data generated with GPT-3.5 \citep{ouyang-etal-2022-training}.

\subsection{Model Compression}
\label{sec:model-compression}

As it is not possible to experiment with every existing model compression approach, we select four of the most widely adopted methods. We report the implementation details in Appendix \ref{app:infrastructure} and complete hyperparameter selection in Appendix \ref{app:hyperparameters}.

\paragraph{Quantization.}

For quantization, we trial \textbf{AWQ} \citep{lin-etal-2024-awq} and \textbf{GPTQ} \citep{frantar-etal-2023-optq}. In both cases, we use 4-bit weight quantization, which sees minimal performance degradation while enabling efficient inference \citep{frantar-etal-2024-marlin}.

\paragraph{Pruning.}

For pruning, we employ \textbf{SparseGPT} \citep{frantar-alistarh-2023-sparsegpt} and \textbf{Wanda} \citep{sun-etal-2024-simple}. In both cases, we focus on the 2:4 semi-structured (50\%) sparsity setting, which enables inference speedups on GPUs \citep{mishra-etal-2021-accelerating}.

\subsection{Evaluation Tasks}
\label{sec:evaluation-tasks}

To offer an impartial selection of evaluation tasks, we adopt all zero-shot tasks used in the original work to evaluate AWQ, GPTQ, SparseGPT, and Wanda. Namely, ARC (easy and challenge sets) \citep{clark-etal-2018-think}, BoolQ \citep{clark-etal-2019-boolq}, HellaSwag \citep{zellers-etal-2019-hellaswag}, LAMBADA \citep{paperno-etal-2016-lambada}, OpenBookQA \citep{banerjee-etal-2019-careful}, PIQA \citep{bisk-etal-2020-piqa}, RTE \citep{dagan-etal-2006-pascal}, StoryCloze \citep{mostafazadeh-etal-2016-corpus}, and WinoGrande \citep{sakaguchi-etal-2021-winogrande}.

\section{Results}

\begin{figure*}[t]
\centering
\includegraphics[scale=0.90]{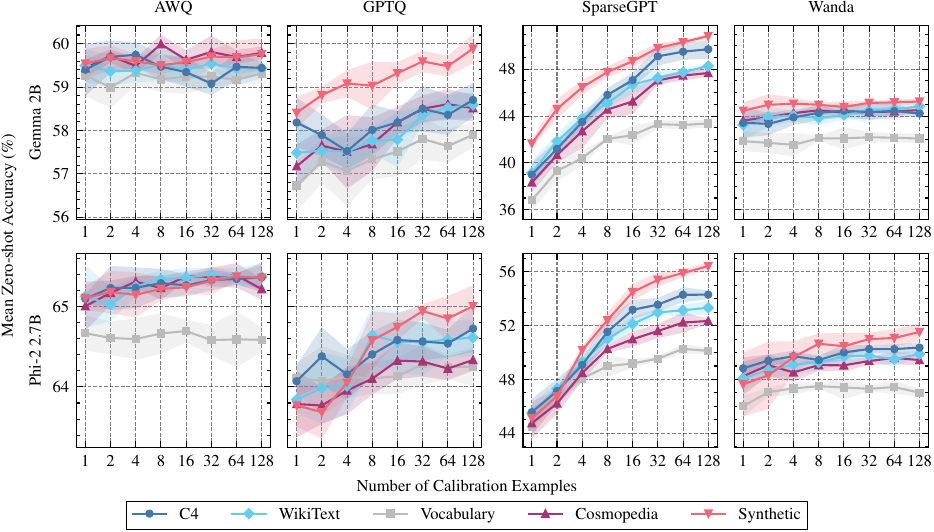}
\caption{The mean zero-shot accuracy when compressing Gemma 2B and Phi-2 with each method. We present the mean value and standard deviation (shaded) across five distinct calibration sets sampled from each data source.}
\label{fig:data-quantity-ablation}
\end{figure*}

\begin{figure*}[t]
\centering
\includegraphics[scale=0.95]{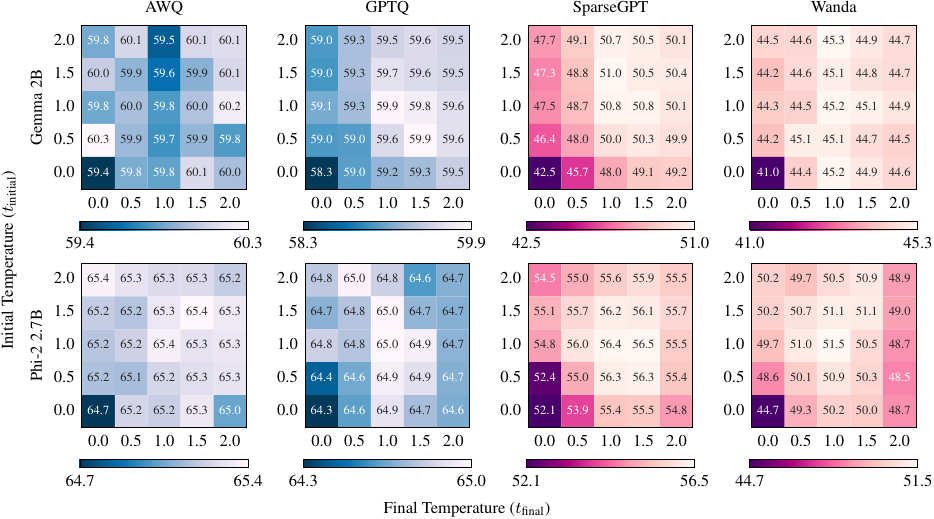}
\caption{A joint parameter search for $t_\text{initial}$ and $t_\text{final}$ using $n=10$ (\S \ref{sec:synthesizing-calibration-data}) with Gemma 2B and Phi-2. We report the mean task accuracy across five distinct calibration sets.}
\label{fig:sampling-schedule-ablation}
\end{figure*}

Table \ref{tab:task_results} presents the average performance across all downstream tasks (\S \ref{sec:evaluation-tasks}) for every model tested (\S \ref{sec:models}).\footnote{Complete results are presented in Appendix \ref{app:complete_results}.} For self-calibration, we set $t$\textsubscript{initial} and $t$\textsubscript{final} as 1.0 (i.e. standard sampling), to enable a fair comparison between models. However, we emphasize that the careful selection of these parameters could lead to further performance improvements. We provide a deeper analysis surrounding the impact of the temperature schedule in \S \ref{sec:sampling-strategy-ablation}.

\paragraph{Self-calibration outperforms other synthetic datasets.}

We observe that the performance of self-calibration matches or exceeds other synthetic datasets in 17 out of 20 instances. For example, when quantizing Gemma 2B with GPTQ, self-calibration records a mean accuracy of 59.9\%, compared to 58.5\% with Cosmopedia and 57.9\% with Vocabulary. Similarly, when pruning Llama 3.1 8B with SparseGPT, self-calibration offers a 2.9 point increase in mean accuracy compared to Cosmopedia (53.8\% versus 50.9\%). This suggests that self-calibration may produce calibration data that is more representative of the training distribution of each model, compared to other synthetic datasets.

\paragraph{Self-calibration can outperform real-world data.}

Our results show that for Phi-2, Gemma 2B, and OPT 6.7B, self-calibration achieves the highest mean accuracy compared to all other datasets in all but one instance. The only exception is when pruning OPT 6.7B with SparseGPT, where self-calibration ranks second to C4 (52.7\% with self-calibration versus 52.8\% with C4). Although self-calibration does not outperform real data for Mistral 7B and Llama 3.1 8B, we observe that the performance is as competitive with real data as Cosmopedia (i.e. matches or outperforms Cosmopedia in five out eight instances). These outcomes suggest that using self-calibration for model compression results in downstream performance that is at least comparable to that of real data.

\paragraph{Pruning benefits the most from self-calibration.}

Across all models and both pruning methods, self-calibration results in higher mean accuracy compared to other synthetic data. For example, when pruning Llama 3.1 8B with Wanda, self-calibration is second only to WikiText by a 0.1 point difference (49.1\% compared to 49.2\% with WikiText) whilst also being 1.4 points higher than Cosmopedia. We also observe that quantization methods appear less sensitive to the calibration data. For example, the difference between the best and worst performing calibration data source for Gemma 2B is 0.6\% with AWQ and 2.0\% with GPTQ. In contrast, there is a range of 7.5\% with SparseGPT and 3.2\% with Wanda. This suggests that the choice of calibration dataset is less critical when applying quantization to language models, corroborating earlier findings from \citet{williams-aletras-2024-impact}.

\paragraph{Random vocabulary consistently underachieves.}

For every model and compression method, we observe that random calibration data (i.e. Vocabulary) produces the lowest performance. In comparison to C4, compressing Phi-2 with this random synthetic calibration data degrades performance by 0.9\% for AWQ, 0.5\% for GPTQ, 4.2\% for SparseGPT, and 3.4\% for Wanda. This illustrates that purely random synthetic data is suboptimal for calibration, even for quantization which may be less sensitive.

\section{Analysis}

\subsection{Calibration Data Quantity Ablation}
\label{sec:data-quantity-ablation}

\paragraph{Methodology.}

To assess how the quantity of calibration data impacts performance, we experiment with calibration sets of different sizes. For each calibration set, we trial subsets of $n$ examples, where $n \in \{1, 2, 4, 8, 16, 32, 64, 128\}$. We repeat this process across five distinct calibration sets sampled from each source of calibration data.\footnote{We perform this ablation using smaller models (Gemma 2B and Phi-2) due to computational resource constraints.} 

\paragraph{Self-calibration may be more sample efficient.}

In the case of pruning, self-calibration may offer comparable or greater performance with less data. For example, when pruning Phi-2 with SparseGPT, C4 reaches a mean accuracy of 54.3\% with 128 examples, while self-calibration achieves 54.5\% with only 16 examples. While the same trend is visible for GPTQ, the performance margin between data sources is too small to draw the same conclusion. Finally, we note that improved sample efficiency can reduce the computational cost of model compression \citep{frantar-alistarh-2023-sparsegpt}. In practical terms, this can enable (1) fewer forward passes, as a direct result of fewer examples, or (2) an increased batch size, due to fewer intermediate activations.

\subsection{Sampling Strategy Ablation}
\label{sec:sampling-strategy-ablation}

\paragraph{Methodology.}

To investigate how the parameters of our sampling strategy (\S  \ref{sec:synthesizing-calibration-data}) impact performance, we explore a broad range of values: $t_\text{initial},\, t_\text{final} \in \{0.0, 0.5, 1.0, 1.5, 2.0\}$. We emphasize that certain subsets of these values are equivalent to several standard decoding strategies:

\begin{itemize}[left=0pt]
\item \textbf{Greedy decoding.} When both $t_\text{initial} = 0$ and $t_\text{final} = 0$, this is equivalent to selecting the token with the highest probability at every timestep.

\item \textbf{Standard sampling.} Using a combination of $t_\text{initial} = 1$ and $t_\text{final} = 1$ is equivalent to applying softmax without a temperature parameter.

\item \textbf{Temperature sampling.} When $t_\text{initial} = t_\text{final}$, a constant temperature is maintained throughout generation, equivalent to temperature sampling.
\end{itemize}

\paragraph{Sampling strategy can influence performance.}

Figure \ref{fig:sampling-schedule-ablation} presents the influence of the sampling strategy parameters upon mean task accuracy. For SparseGPT and Wanda, the careful selection of sampling parameters may offer improved performance. For example, Gemma 2B sees slightly elevated performance when using a higher initial temperature and moderate final temperature. Conversely, using both a low initial and final temperature leads to substantially lower performance.

\paragraph{Selecting sampling parameters is not essential.}

We observe that it is possible to achieve within 0.5 points of the maximum performance through using only standard sampling (i.e. $t_\text{initial}=t_\text{final}=1$). This suggests that self-calibration can achieve reasonable performance with little attention towards the specific parameters used. Consequently, we suspect that using the model itself to generate calibration data is a relatively stable and reliable method.

\subsection{Calibration Data Analysis}
\label{sec:calibration-data-analysis}

\paragraph{Methodology.}

The content and style of text can vary markedly between calibration data sources. Consequently, we seek to analyze how the text characteristics differ between them. To this end, we employ a variety of automatic metrics to assess various text characteristics of the calibration sets.

\begin{itemize}[left=0pt]
\item \textbf{Perplexity.} As an indirect indicator of text quality, we calculate the average perplexity across examples in the calibration set for a given model.

\item \textbf{Repetitions.} Following \citet{welleck-etal-2020-neural}, we report the average fraction of repeated tokens per sequence. More formally, this is computed across each sequence $w$ of length $L$ in dataset $\mathcal{D}$, where $\mathbb{I}$ denotes the binary indicator function: $$R=\frac{1}{|\mathcal{D}|L}\sum_{w \in \mathcal{D}}\sum_{i=1}^L \mathbb{I}(w_i \in w_{1:i-1})$$

\item \textbf{Vocabulary coverage.} To assess the lexical diversity of the calibration sets, we report the vocabulary coverage. We define this as the ratio between the subword tokens present in the calibration set and in the model vocabulary.

\item \textbf{N-gram diversity.} Following \citet{meister-etal-2023-locally}, we report the average fraction of unique $n$-grams ($n \in \{1,2,3,4\}$) in the calibration set: $$D=\frac{1}{N}\sum_{n=1}^N\frac{\text{\# unique } n\text{-grams}}{\text{\# total } n\text{-grams}}$$

\item \textbf{Zipf's coefficient.} Finally, we examine the extent to which the calibration set follows Zipf's law. Specifically, we calculate the fit of the exponent corresponding to the calibration set. Natural language text tends to have a value close to one.
\end{itemize}

\paragraph{Self-calibration data is generally coherent text.}

Table \ref{tab:self_calibration_examples_main} presents self-calibration data from Gemma 2 and Llama 3.1 8B. For brevity, we select the first three texts generated by each model. We observe that the self-generated text is typically coherent and fluent in both models. Moreover, the content is routinely semantically plausible. These properties are somewhat supported by the perplexity results in Table \ref{tab:results_data_main}, with self-calibration demonstrating substantially lower perplexity than real data.

\paragraph{Self-calibration may produce less diverse text.}

Table \ref{tab:results_data_main} presents the text characteristics for Gemma 2B and Llama 3.1 8B across all datasets.\footnote{We observe similar results in other models (Appendix \ref{app:calibration-data-analysis}).} Compared to real data sources (i.e. C4 and WikiText), self-calibration data differs across various metrics. For example, self-calibration data from Llama 3.1 8B has a lower vocabulary coverage (0.15 versus 0.16-0.18) and $n$-gram diversity (0.58 versus 0.62-0.65). However, self-calibration data has a higher Zipf's coefficient (1.24 versus 1.12-1.16) and slightly elevated repetitions (0.66 versus 0.64-0.65). Overall, this suggests that self-calibration data is typically less diverse compared to real data.

\begin{table}[t]
\scriptsize
\centering
\begin{tabular}{rp{6.7cm}}
\toprule
\# & Generated Text \\
\midrule
\multicolumn{2}{c}{Gemma 2B} \\
\midrule
1 & \specialtoken{<bos>}The <b>G36S<\/b> is an assault rifle created for the German Army from 1997 to 2010 by Heckler \& Koch. It is a simplified... \\
\cdashlinelr{1-2}
2 & \specialtoken{<bos>}Are you considering making an investment in a used or new Mercedes-Benz S-Class? Make Mercedes-Benz of Houston your... \\
\cdashlinelr{1-2}
3 & \specialtoken{<bos>}I recently created a poll to see what everyone thinks the best of the current generation of S13's are. I have gotten some great... \\
\midrule
\multicolumn{2}{c}{Llama 3.1 8B} \\
\midrule
1 & \specialtoken{<|begin\_of\_text|>}You are at:Home\guillemotright Lifestyle\guillemotright Food\guillemotright I have a problem... and it's called peanut butter!\visiblenewline I have a problem... and it's... \\
\cdashlinelr{1-2}
2 & \specialtoken{<|begin\_of\_text|>}When we're in the heat of our journey, when our plans and goals and hopes and dreams and desires are in control... \\
\cdashlinelr{1-2}
3 & \specialtoken{<|begin\_of\_text|>}This article by David K. Li from NBC News on February 9, 2021, talks about the increase of the vaccine mandate... \\
\bottomrule
\end{tabular}
\caption{The starting segment of the first three synthetic texts generated by Gemma 2B and Llama 3.1 8B, using standard sampling.}
\label{tab:self_calibration_examples_main}
\end{table}

\begin{table}[t]
\scriptsize
\centering
\begingroup
\addtolength{\tabcolsep}{-1.5pt}
\begin{tabular}{lrrrrr}
\toprule
Dataset & \multicolumn{1}{c}{PPL} & \multicolumn{1}{c}{Rep.} & \multicolumn{1}{c}{Cov.} & \multicolumn{1}{c}{Div.} & \multicolumn{1}{c}{Zipf} \\
\midrule 
\multicolumn{6}{c}{Gemma 2B} \\
\midrule
C4 & \result{19.30}{1.06} & \result{0.66}{0.01} & \result{0.10}{0.00} & \result{0.63}{0.01} & \result{1.16}{0.01} \\
WikiText & \result{14.93}{0.58} & \result{0.68}{0.00} & \result{0.09}{0.00} & \result{0.65}{0.00} & \result{1.12}{0.01} \\
\cdashlinelr{1-6}
Vocabulary & 4.31$\times$10\textsuperscript{6} & \result{0.00}{0.00} & \result{0.64}{0.00} & \result{0.96}{0.00} & \result{0.27}{0.00} \\
Cosmopedia & \result{6.49}{0.22} & \result{0.59}{0.01} & \result{0.09}{0.00} & \result{0.65}{0.01} & \result{1.19}{0.01} \\
Self-calibration & \result{7.22}{0.15} & \result{0.68}{0.00} & \result{0.07}{0.00} & \result{0.59}{0.00} & \result{1.25}{0.01} \\
\midrule
\multicolumn{6}{c}{Llama 3.1 8B} \\
\midrule
C4 & \result{8.65}{0.50} & \result{0.64}{0.00} & \result{0.18}{0.00} & \result{0.62}{0.01} & \result{1.16}{0.02} \\
WikiText & \result{6.75}{0.11} & \result{0.65}{0.00} & \result{0.16}{0.00} & \result{0.65}{0.00} & \result{1.12}{0.01} \\
\cdashlinelr{1-6}
Vocabulary & 7.61$\times$10\textsuperscript{5} & \result{0.01}{0.00} & \result{0.87}{0.00} & \result{0.91}{0.00} & \result{0.49}{0.00} \\
Cosmopedia & \result{3.37}{0.16} & \result{0.55}{0.02} & \result{0.18}{0.01} & \result{0.65}{0.01} & \result{1.17}{0.01} \\
Self-calibration & \result{6.29}{0.09} & \result{0.66}{0.00} & \result{0.15}{0.00} & \result{0.58}{0.00} & \result{1.24}{0.00} \\
\bottomrule
\end{tabular}
\endgroup
\caption{Text characteristics across all calibration sets for Gemma 2B and Llama 3.1 8B, with standard deviation denoted in subscript.}
\label{tab:results_data_main}
\end{table}

\section{Conclusion}

In this paper, we proposed self-calibration for LLM quantization and pruning as a solution to mitigate concerns about the availability, quality, and representativeness of training data. Our proposed approach is intuitive and requires no external data sources, instead relying on the model itself. We empirically demonstrated that self-calibration maintains comparable or greater downstream task performance across a variety of models and compression methods. Surprisingly, our results also revealed that self-calibration can enable higher downstream task performance than using real data. We hope that our study will inspire further work on the application of synthetic data to LLM compression.

\section*{Limitations}

In this study, we experimented with English models and evaluation tasks, and therefore only English calibration data. However, recent work has illustrated the importance of language-specific calibration data when compressing multilingual models \citep{zeng-etal-2024-multilingual, kurz-etal-2024-investigating}. Although we anticipate that our approach will generalize to multilingual models, we hope to explore this matter further in a future work.

\section*{Ethical Considerations}

Language models are capable of generating text that is incorrect, biased, and harmful \citep{weidinger-etal-2022-taxonomy}. To compress a given model, our approach requires the unsupervised generation of calibration data from the model itself. Consequently, the calibration data may contain material that is problematic. However, we note that this is unlikely to introduce new safety issues in the compressed model. For the generated calibration data to contain problematic content, it must have already been encoded in the weights of the original model.

\section*{Acknowledgments}

We are grateful to Vladimir Poroshin, Vitor Jeronymo, Szymon Palucha, Christopher May, Mario Sanger, and the anonymous reviewers for their invaluable feedback. MW is supported by the Centre for Doctoral Training in Speech and Language Technologies (SLT) and their Applications funded by UK Research and Innovation grant EP/S023062/1. NA is supported by EPSRC grant EP/Y009800/1, part of the RAI UK Keystone projects.

\bibliography{anthology, custom}

\clearpage
\appendix

\section{Infrastructure}
\label{app:infrastructure}

We use the model implementations and prepared datasets from the Hugging Face Transformers \citep{wolf-etal-2020-transformers} and Datasets \citep{lhoest-etal-2021-datasets} libraries, respectively. For pruning with SparseGPT and Wanda, we adopt the implementation from \citet{sun-etal-2024-simple}. For quantization with AWQ and GPTQ, we use the NVIDIA TensorRT Model Optimizer and AutoGPTQ libraries, respectively.\footnote{See \url{https://nvidia.github.io/TensorRT-Model-Optimizer} and \url{https://github.com/AutoGPTQ/AutoGPTQ}.} To enable reproducible model evaluations, we use the EleutherAI Language Model Evaluation Harness \citep{gao-etal-2023-framework}. All experiments are conducted using a single NVIDIA A100 80GB GPU.

\section{Evaluation Datasets}
\label{app:evaluation-datasets}

Table \ref{tab:evaluation_statistics} lists the number of examples used from the relevant dataset split in each evaluation task. This is either the validation or test split, as implemented by \citet{gao-etal-2023-framework}.

\section{Hyperparameters}
\label{app:hyperparameters}

Table \ref{tab:hyperparameters} presents the hyperparameters used in all experiments. For SparseGPT and Wanda, we adopt the hyperparameters used in the original work. For AWQ and GPTQ, we use the hyperparameters from the respective implementations, NVIDIA TensorRT Model Optimizer and AutoGPTQ (\S \ref{app:infrastructure}).

\section{Calibration Data Analysis}
\label{app:calibration-data-analysis}

Supplementary to the text characteristic results for Gemma 2 and Llama 3.1 8B presented in \S \ref{sec:calibration-data-analysis}, we present the results for Phi-2 2.7B, OPT 6.7B, and Mistral 7B in Table \ref{tab:results_data_supplementary}. Finally, we also present self-calibration examples for the remaining models (Phi-2 2.7B, OPT 6.7B, and Mistral 7B) in Table \ref{tab:self_calibration_examples_supplementary}.

\section{Complete Results}
\label{app:complete_results}

In addition to the summarized results (Table \ref{tab:task_results}, we present the task performance across compression methods and calibration data sources for each model: Gemma 2B (Table \ref{tab:results_tasks_gemma_2b}), Phi-2 2.7B (Table \ref{tab:results_tasks_phi_2}), OPT 6.7B (Table \ref{tab:results_tasks_opt_6.7b}), Mistral 7B (Table \ref{tab:results_tasks_mistral_7b_v0.3}), and Llama 3.1 8B (Table \ref{tab:results_tasks_llama_3.1_8b}).

\begin{table}[t]
\small
\centering
\begin{tabular}{lr}
\toprule
Dataset & \#
Examples \\
\midrule
ARC-Easy \citep{clark-etal-2018-think} & 2,376 \\
ARC-Challenge \citep{clark-etal-2018-think} & 1,172 \\
BoolQ \citep{clark-etal-2019-boolq} & 3,270 \\
HellaSwag \citep{zellers-etal-2019-hellaswag} & 10,042 \\
LAMBADA \citep{paperno-etal-2016-lambada} & 5,153 \\
OpenBookQA \citep{banerjee-etal-2019-careful} & 500 \\
PIQA \citep{bisk-etal-2020-piqa} & 1,838 \\
RTE \citep{dagan-etal-2006-pascal} & 277 \\
StoryCloze \citep{mostafazadeh-etal-2016-corpus} & 1,511 \\
WinoGrande \citep{sakaguchi-etal-2021-winogrande} & 1,267 \\
\bottomrule
\end{tabular}
\caption{Number of examples in each evaluation task.}
\label{tab:evaluation_statistics}
\end{table}

\begin{table}[t]
\small
\centering
\begin{tabular}{llc}
\toprule
Method & Hyperparameter & Value \\
\midrule
\multirow{6}{*}{AWQ} & Bits per Weight & 4 \\
& Clip Step Size & 0.05 \\
& Group Size & 128 \\
& Maximum Clip Tokens & 64 \\
& Minimum Clip Ratio & 0.5 \\
& Scale Step Size & 0.1 \\
\midrule
\multirow{6}{*}{GPTQ}  & Bits per Weight & 4 \\
 & Dampening & 0.01 \\
 & Descending Activation Order & Yes \\
 & Group Size & 128 \\
 & Symmetric Quantization & Yes \\
 & True Sequential Quantization & Yes \\
\midrule
 \multirow{3}{*}{SparseGPT} & Dampening & 0.01 \\
 & Group Size & 128 \\
 & Sparsity & 2:4 \\
\midrule
\multirow{2}{*}{Wanda} & Group Size & 1 \\
 & Sparsity & 2:4 \\
\bottomrule
\end{tabular}
\caption{The hyperparameters used in all experiments.}
\label{tab:hyperparameters}
\end{table}

\begin{table}[t]
\scriptsize
\centering
\begin{tabular}{rp{6.7cm}}
\toprule
\# & Generated Text \\
\midrule
\multicolumn{2}{c}{Phi-2 2.7B} \\
\midrule
1 & \specialtoken{<|endoftext|>}\visiblenewline\visiblenewline\visiblenewline def simple\_math\_problem() -> int:\visiblenewline    \textquotesingle\textquotesingle\textquotesingle\visiblenewline    Nathan collects all the leaves from his 8 bushes.\visiblenewline    Each bush has 16 plants... \\
\cdashlinelr{1-2}
2 & \specialtoken{<|endoftext|>}\visiblenewline\visiblenewline \#\# TAKING OWNERSHIP OF WORKPLACE FEARS \visiblenewline\visiblenewline Good morning everyone,\visiblenewline\visiblenewline I'm here today to talk... \\
\cdashlinelr{1-2}
3 & \specialtoken{<|endoftext|>}\visiblenewline\visiblenewline \#\# BOOSTING ANIMAL POPULATION IN INDIA\visiblenewline\visiblenewline India is known for its rich biodiversity, with a variety of... \\
\midrule
\multicolumn{2}{c}{OPT 6.7B} \\
\midrule
1 & \specialtoken{<\/s>}It's an interesting concept, but there's no way anyone can get past the cost. I can't see this going anywhere.\visiblenewline Well this is what's... \\
\cdashlinelr{1-2}
2 & \specialtoken{<\/s>}You are here\visiblenewline\visiblenewline Olympics Day 10: US men, Phelps, Lochte \& swimming's greatest\visiblenewline\visiblenewline Updated: Wednesday, 20 August 2014... \\
\cdashlinelr{1-2}
3 & \specialtoken{<\/s>}Tampa Bay Lightning\visiblenewline I'm a simple man, I see Lightning, I read Stamkos.\visiblenewline Yeah, the Bolts are gonna be so fun to watch next year... \\
\midrule
\multicolumn{2}{c}{Mistral 7B} \\
\midrule
1 & \specialtoken{<s>}What with the heat of the summer and a seemingly endless amount of time spent outside in awe at the scenery and the local wildlife... \\
\cdashlinelr{1-2}
2 & \specialtoken{<s>}While working on an assignment on how to manage a conflict in our teams at University, I was inspired to do so in my own work... \\
\cdashlinelr{1-2}
3 & \specialtoken{<s>}Using ``the old reliable'' ``the old faithful'' methods of lead generation can quickly become...  well, let's say, repetitive, uninspired...   \\
\bottomrule
\end{tabular}
\caption{The starting segment of the first three synthetic texts generated by Phi-2 2.7B, OPT 6.7B, and Mistral 7B, using standard sampling.}
\label{tab:self_calibration_examples_supplementary}
\end{table}

\begin{table}[t]
\scriptsize
\centering
\begingroup
\addtolength{\tabcolsep}{-1.5pt}
\begin{tabular}{lrrrrrr}
\toprule
Dataset & \multicolumn{1}{c}{PPL} & \multicolumn{1}{c}{Rep.} & \multicolumn{1}{c}{Cov.} & \multicolumn{1}{c}{Div.} & \multicolumn{1}{c}{Zipf} \\
\midrule
\multicolumn{6}{c}{Phi-2 2.7B} \\
\midrule
C4 & \result{13.01}{0.59} & \result{0.65}{0.01} & \result{0.44}{0.01} & \result{0.63}{0.01} & \result{1.16}{0.02} \\
WikiText & \result{10.32}{0.10} & \result{0.65}{0.00} & \result{0.40}{0.01} & \result{0.65}{0.00} & \result{1.12}{0.01} \\
\cdashlinelr{1-6}
Vocabulary & 1.98$\times$10\textsuperscript{5} & \result{0.02}{0.00} & \result{0.99}{0.00} & \result{0.87}{0.00} & \result{0.66}{0.00} \\
Cosmopedia & \result{4.32}{0.08} & \result{0.59}{0.01} & \result{0.40}{0.01} & \result{0.65}{0.00} & \result{1.16}{0.01} \\
Self-calibration & \result{2.41}{0.02} & \result{0.67}{0.00} & \result{0.33}{0.00} & \result{0.57}{0.00} & \result{1.24}{0.00} \\
\midrule
\multicolumn{6}{c}{OPT 6.7B} \\
\midrule
C4 & \result{11.80}{0.46} & \result{0.65}{0.01} & \result{0.44}{0.01} & \result{0.63}{0.01} & \result{1.16}{0.01} \\
WikiText & \result{11.05}{0.16} & \result{0.65}{0.00} & \result{0.40}{0.01} & \result{0.65}{0.00} & \result{1.12}{0.01} \\
\cdashlinelr{1-6}
Vocabulary & 2.02$\times$10\textsuperscript{5} & \result{0.02}{0.00} & \result{0.99}{0.00} & \result{0.87}{0.00} & \result{0.66}{0.00} \\
Cosmopedia & \result{5.76}{0.13} & \result{0.60}{0.01} & \result{0.40}{0.01} & \result{0.65}{0.01} & \result{1.15}{0.01} \\
Self-calibration & \result{7.06}{0.21} & \result{0.66}{0.00} & \result{0.32}{0.01} & \result{0.57}{0.01} & \result{1.25}{0.01} \\
\midrule
\multicolumn{6}{c}{Mistral 7B} \\
\midrule
C4 & \result{7.81}{0.17} & \result{0.65}{0.01} & \result{0.47}{0.01} & \result{0.63}{0.00} & \result{1.15}{0.01} \\
WikiText & \result{5.81}{0.07} & \result{0.67}{0.00} & \result{0.42}{0.01} & \result{0.65}{0.00} & \result{1.10}{0.01} \\
\cdashlinelr{1-6}
Vocabulary & 1.64$\times$10\textsuperscript{5} & \result{0.03}{0.00} & \result{0.98}{0.00} & \result{0.89}{0.00} & \result{0.55}{0.00} \\
Cosmopedia & \result{3.07}{0.03} & \result{0.53}{0.01} & \result{0.46}{0.00} & \result{0.67}{0.01} & \result{1.15}{0.01} \\
Self-calibration & \result{5.79}{0.15} & \result{0.66}{0.00} & \result{0.41}{0.00} & \result{0.59}{0.00} & \result{1.24}{0.00} \\
\bottomrule
\end{tabular}
\endgroup
\caption{Text characteristics across all calibration sets for Phi-2 2.7B, OPT 6.7B, and Mistral 7B, with standard deviation denoted in subscript.}
\label{tab:results_data_supplementary}
\end{table}

\begin{table*}[t]
\scriptsize
\centering
\begin{tabular}{llrrrrrrrrrrr}
\toprule
Method & Dataset & \multicolumn{1}{c}{ARC-e} & \multicolumn{1}{c}{ARC-c} & \multicolumn{1}{c}{BoolQ} & \multicolumn{1}{c}{HS} & \multicolumn{1}{c}{LMBD} & \multicolumn{1}{c}{OBQA} & \multicolumn{1}{c}{PIQA} & \multicolumn{1}{c}{RTE} & \multicolumn{1}{c}{SC} & \multicolumn{1}{c}{WG} & \multicolumn{1}{c}{Mean} \\
\midrule
- & - & \multicolumn{1}{c}{74.1} & \multicolumn{1}{c}{40.4} & \multicolumn{1}{c}{69.7} & \multicolumn{1}{c}{52.8} & \multicolumn{1}{c}{58.3} & \multicolumn{1}{c}{30.8} & \multicolumn{1}{c}{77.2} & \multicolumn{1}{c}{64.3} & \multicolumn{1}{c}{74.9} & \multicolumn{1}{c}{64.6} & \multicolumn{1}{c}{60.7} \\
\cmidrule{1-13}
\multirow[c]{5}{*}{AWQ} & C4 & \result{73.4}{0.1} & \result{39.4}{0.6} & \result{67.8}{0.3} & \result{51.4}{0.1} & \result{59.6}{1.1} & \result{29.2}{0.6} & \result{76.8}{0.1} & \result{59.1}{2.3} & \result{74.0}{0.4} & \result{64.1}{0.4} & \result{59.5}{0.2} \\
 & WikiText & \result{73.4}{0.4} & \result{39.5}{0.8} & \result{68.0}{1.5} & \result{51.5}{0.0} & \result{59.1}{1.2} & \result{29.4}{0.6} & \result{76.6}{0.3} & \result{58.6}{1.7} & \result{74.0}{0.3} & \result{64.7}{0.5} & \result{59.5}{0.2} \\
\cdashlinelr{2-13}
 & Vocabulary & \result{72.1}{0.3} & \result{39.6}{0.4} & \result{67.1}{1.0} & \result{51.2}{0.1} & \result{57.3}{0.1} & \result{29.0}{0.8} & \result{75.9}{0.5} & \result{61.2}{1.3} & \result{74.5}{0.2} & \result{64.5}{0.2} & \result{59.3}{0.2} \\
 & Cosmopedia & \result{73.4}{0.4} & \result{39.5}{0.6} & \result{68.3}{0.9} & \result{51.3}{0.1} & \result{60.5}{0.8} & \result{29.4}{1.0} & \result{76.8}{0.2} & \result{60.0}{1.5} & \result{74.1}{0.3} & \result{64.7}{0.5} & \result{59.8}{0.2} \\
 & Self-calibration & \result{73.3}{0.2} & \result{40.6}{0.4} & \result{68.6}{0.2} & \result{51.9}{0.2} & \result{59.5}{0.6} & \result{28.8}{0.2} & \result{76.6}{0.3} & \result{60.5}{3.0} & \result{74.1}{0.3} & \result{63.6}{0.3} & \result{59.8}{0.4} \\
\cmidrule{1-13}
\multirow[c]{5}{*}{GPTQ} & C4 & \result{72.8}{0.7} & \result{38.4}{0.8} & \result{68.7}{1.4} & \result{50.9}{0.2} & \result{54.9}{1.7} & \result{27.8}{1.1} & \result{76.1}{0.3} & \result{60.6}{2.8} & \result{73.6}{0.4} & \result{63.1}{0.7} & \result{58.7}{0.4} \\
 & WikiText & \result{71.3}{0.8} & \result{37.2}{0.4} & \result{67.3}{0.8} & \result{51.3}{0.2} & \result{55.3}{0.3} & \result{29.0}{0.9} & \result{76.0}{0.4} & \result{61.8}{1.8} & \result{73.4}{0.4} & \result{63.4}{0.4} & \result{58.6}{0.3} \\
\cdashlinelr{2-13}
 & Vocabulary & \result{70.8}{1.0} & \result{36.3}{0.7} & \result{69.1}{0.7} & \result{50.3}{0.3} & \result{53.2}{1.8} & \result{29.1}{0.9} & \result{75.9}{0.3} & \result{58.3}{1.4} & \result{72.3}{0.5} & \result{63.6}{0.7} & \result{57.9}{0.3} \\
 & Cosmopedia & \result{72.8}{0.9} & \result{38.1}{0.5} & \result{68.1}{0.6} & \result{51.2}{0.4} & \result{54.0}{1.2} & \result{29.2}{2.1} & \result{75.3}{0.6} & \result{59.0}{2.3} & \result{73.9}{0.5} & \result{63.6}{1.0} & \result{58.5}{0.3} \\
 & Self-calibration & \result{73.5}{0.8} & \result{40.0}{0.4} & \result{68.8}{0.8} & \result{52.0}{0.1} & \result{57.6}{1.2} & \result{29.6}{0.7} & \result{76.6}{0.3} & \result{61.7}{2.5} & \result{74.0}{0.5} & \result{65.1}{0.7} & \result{59.9}{0.3} \\
\cmidrule{1-13}
\multirow[c]{5}{*}{SparseGPT} & C4 & \result{60.2}{1.1} & \result{25.9}{1.2} & \result{63.4}{0.9} & \result{39.9}{0.3} & \result{39.7}{2.2} & \result{21.3}{1.2} & \result{70.0}{0.3} & \result{55.7}{2.9} & \result{64.0}{0.3} & \result{57.0}{1.1} & \result{49.7}{0.8} \\
 & WikiText & \result{58.0}{0.9} & \result{25.5}{0.6} & \result{62.8}{0.6} & \result{37.6}{0.1} & \result{37.0}{1.4} & \result{20.8}{0.9} & \result{67.2}{0.5} & \result{55.7}{0.5} & \result{62.3}{0.5} & \result{55.8}{1.0} & \result{48.3}{0.2} \\
\cdashlinelr{2-13}
 & Vocabulary & \result{52.0}{0.9} & \result{21.1}{0.5} & \result{61.8}{0.4} & \result{33.5}{0.1} & \result{16.5}{0.4} & \result{18.1}{0.9} & \result{66.5}{0.6} & \result{53.0}{0.5} & \result{56.5}{0.3} & \result{54.6}{1.1} & \result{43.4}{0.3} \\
 & Cosmopedia & \result{60.1}{1.0} & \result{25.7}{0.7} & \result{62.2}{0.1} & \result{37.8}{0.3} & \result{29.6}{1.4} & \result{19.8}{0.9} & \result{68.1}{0.5} & \result{56.3}{1.9} & \result{61.2}{0.6} & \result{55.9}{0.9} & \result{47.7}{0.3} \\
 & Self-calibration & \result{63.0}{0.5} & \result{28.0}{0.8} & \result{62.7}{0.3} & \result{40.5}{0.2} & \result{38.1}{1.2} & \result{22.1}{0.5} & \result{70.7}{0.7} & \result{57.5}{1.5} & \result{66.7}{0.2} & \result{59.0}{0.7} & \result{50.8}{0.2} \\
\cmidrule{1-13}
\multirow[c]{5}{*}{Wanda} & C4 & \result{54.9}{0.6} & \result{24.6}{0.6} & \result{53.7}{2.8} & \result{36.4}{0.1} & \result{19.8}{0.3} & \result{17.0}{0.2} & \result{66.5}{0.5} & \result{54.6}{1.6} & \result{59.1}{0.3} & \result{55.5}{0.3} & \result{44.2}{0.2} \\
 & WikiText & \result{54.4}{0.5} & \result{23.9}{0.5} & \result{60.6}{1.4} & \result{35.8}{0.2} & \result{20.2}{0.9} & \result{17.2}{0.9} & \result{66.4}{0.2} & \result{55.6}{1.4} & \result{58.9}{0.3} & \result{55.4}{0.5} & \result{44.8}{0.4} \\
\cdashlinelr{2-13}
 & Vocabulary & \result{51.0}{0.4} & \result{22.1}{0.3} & \result{54.4}{2.7} & \result{33.8}{0.1} & \result{13.4}{0.4} & \result{16.3}{1.1} & \result{65.8}{0.2} & \result{51.6}{1.8} & \result{57.7}{0.3} & \result{54.7}{0.6} & \result{42.1}{0.4} \\
 & Cosmopedia & \result{54.4}{0.8} & \result{24.4}{0.6} & \result{62.1}{0.2} & \result{35.8}{0.2} & \result{16.6}{0.3} & \result{16.8}{0.6} & \result{66.4}{0.5} & \result{55.1}{0.5} & \result{57.5}{0.3} & \result{55.9}{0.6} & \result{44.5}{0.2} \\
 & Self-calibration & \result{56.4}{0.2} & \result{25.6}{0.4} & \result{51.8}{1.4} & \result{37.2}{0.1} & \result{23.1}{0.3} & \result{19.6}{0.5} & \result{67.5}{0.4} & \result{53.9}{1.0} & \result{61.2}{0.3} & \result{56.1}{0.6} & \result{45.2}{0.3} \\
\bottomrule
\end{tabular}
\caption{Task accuracy across five calibration sets for Gemma 2B, with standard deviation denoted in subscript.}
\label{tab:results_tasks_gemma_2b}
\end{table*}

\begin{table*}[t]
\scriptsize
\centering
\begin{tabular}{llrrrrrrrrrrr}
\toprule
Method & Dataset & \multicolumn{1}{c}{ARC-e} & \multicolumn{1}{c}{ARC-c} & \multicolumn{1}{c}{BoolQ} & \multicolumn{1}{c}{HS} & \multicolumn{1}{c}{LMBD} & \multicolumn{1}{c}{OBQA} & \multicolumn{1}{c}{PIQA} & \multicolumn{1}{c}{RTE} & \multicolumn{1}{c}{SC} & \multicolumn{1}{c}{WG} & \multicolumn{1}{c}{Mean} \\
\midrule
- & - & \multicolumn{1}{c}{79.8} & \multicolumn{1}{c}{53.0} & \multicolumn{1}{c}{83.4} & \multicolumn{1}{c}{55.8} & \multicolumn{1}{c}{49.8} & \multicolumn{1}{c}{40.2} & \multicolumn{1}{c}{78.6} & \multicolumn{1}{c}{62.5} & \multicolumn{1}{c}{79.3} & \multicolumn{1}{c}{75.8} & \multicolumn{1}{c}{65.8} \\
\midrule
\multirow[c]{5}{*}{AWQ} & C4 & \result{80.2}{0.3} & \result{51.7}{0.4} & \result{82.3}{0.2} & \result{54.8}{0.1} & \result{47.6}{0.4} & \result{39.8}{0.6} & \result{78.9}{0.3} & \result{65.5}{0.7} & \result{77.7}{0.3} & \result{75.8}{0.6} & \result{65.4}{0.2} \\
 & WikiText & \result{80.4}{0.2} & \result{51.4}{0.7} & \result{83.1}{0.2} & \result{54.6}{0.1} & \result{47.1}{0.5} & \result{39.0}{0.8} & \result{78.9}{0.1} & \result{65.1}{1.2} & \result{77.7}{0.2} & \result{76.2}{0.4} & \result{65.4}{0.2} \\
\cmidrule(lr){2-13}
 & Vocabulary & \result{80.1}{0.2} & \result{50.8}{0.4} & \result{78.7}{0.5} & \result{54.0}{0.1} & \result{45.9}{0.3} & \result{39.5}{0.8} & \result{78.6}{0.4} & \result{65.9}{1.3} & \result{76.6}{0.2} & \result{75.4}{0.7} & \result{64.5}{0.2} \\
 & Cosmopedia & \result{80.2}{0.2} & \result{51.0}{0.5} & \result{81.7}{0.5} & \result{54.7}{0.1} & \result{46.8}{0.1} & \result{39.5}{0.6} & \result{78.3}{0.0} & \result{66.8}{1.4} & \result{77.7}{0.2} & \result{75.8}{0.6} & \result{65.3}{0.2} \\
 & Self-calibration & \result{80.6}{0.3} & \result{51.1}{0.3} & \result{82.9}{0.1} & \result{54.7}{0.1} & \result{47.5}{0.2} & \result{39.3}{0.3} & \result{78.1}{0.2} & \result{65.8}{0.8} & \result{78.1}{0.5} & \result{75.6}{0.7} & \result{65.4}{0.2} \\
\midrule
\multirow[c]{5}{*}{GPTQ} & C4 & \result{79.6}{0.1} & \result{50.9}{0.8} & \result{82.3}{0.7} & \result{54.5}{0.1} & \result{46.8}{0.6} & \result{38.9}{0.7} & \result{78.4}{0.3} & \result{62.1}{1.6} & \result{78.2}{0.4} & \result{75.6}{0.9} & \result{64.7}{0.3} \\
 & WikiText & \result{79.5}{0.3} & \result{50.4}{0.6} & \result{80.8}{0.6} & \result{54.2}{0.1} & \result{46.7}{0.4} & \result{39.1}{0.7} & \result{78.0}{0.6} & \result{64.0}{1.3} & \result{78.0}{0.4} & \result{75.3}{0.6} & \result{64.6}{0.2} \\
\cmidrule(lr){2-13}
 & Vocabulary & \result{79.3}{0.3} & \result{50.3}{0.9} & \result{80.1}{1.5} & \result{53.9}{0.2} & \result{45.6}{0.6} & \result{38.5}{1.6} & \result{77.9}{0.3} & \result{64.1}{1.0} & \result{77.7}{0.2} & \result{75.1}{0.8} & \result{64.3}{0.2} \\
 & Cosmopedia & \result{79.5}{0.4} & \result{50.4}{0.3} & \result{80.9}{1.1} & \result{54.4}{0.1} & \result{45.8}{0.6} & \result{38.6}{0.6} & \result{78.2}{0.5} & \result{63.4}{0.9} & \result{77.5}{0.5} & \result{74.7}{0.8} & \result{64.3}{0.1} \\
 & Self-calibration & \result{79.6}{0.3} & \result{51.6}{0.6} & \result{82.0}{0.7} & \result{54.6}{0.2} & \result{46.9}{0.8} & \result{39.0}{0.4} & \result{77.9}{0.3} & \result{64.5}{0.8} & \result{78.2}{0.6} & \result{75.6}{0.7} & \result{65.0}{0.3} \\
\cmidrule{1-13}
\multirow[c]{5}{*}{SparseGPT} & C4 & \result{69.3}{0.6} & \result{35.1}{0.8} & \result{67.5}{1.1} & \result{42.1}{0.4} & \result{32.6}{0.6} & \result{27.7}{0.9} & \result{72.0}{1.0} & \result{59.2}{2.0} & \result{69.0}{0.3} & \result{68.6}{0.4} & \result{54.3}{0.3} \\
 & WikiText & \result{69.6}{0.7} & \result{35.1}{1.0} & \result{63.1}{0.7} & \result{40.6}{0.3} & \result{33.3}{0.3} & \result{27.4}{0.4} & \result{71.3}{0.7} & \result{57.4}{5.1} & \result{68.2}{0.2} & \result{67.5}{0.8} & \result{53.3}{0.5} \\
\cdashlinelr{2-13}
 & Vocabulary & \result{67.1}{0.3} & \result{32.3}{0.5} & \result{64.4}{0.6} & \result{37.8}{0.1} & \result{20.8}{0.7} & \result{23.3}{0.8} & \result{71.2}{0.5} & \result{57.7}{2.2} & \result{64.0}{0.2} & \result{62.7}{1.3} & \result{50.1}{0.2} \\
 & Cosmopedia & \result{70.5}{1.0} & \result{37.2}{0.8} & \result{64.6}{0.9} & \result{40.6}{0.3} & \result{24.0}{0.5} & \result{27.6}{1.2} & \result{71.2}{0.4} & \result{56.0}{1.9} & \result{66.0}{0.4} & \result{65.8}{0.9} & \result{52.3}{0.2} \\
 & Self-calibration & \result{71.2}{0.3} & \result{37.7}{0.5} & \result{73.4}{1.0} & \result{41.6}{0.3} & \result{31.6}{0.8} & \result{32.1}{0.8} & \result{72.0}{0.5} & \result{65.5}{2.8} & \result{71.0}{0.4} & \result{68.2}{0.5} & \result{56.4}{0.3} \\
\cmidrule{1-13}
\multirow[c]{5}{*}{Wanda} & C4 & \result{68.1}{0.4} & \result{33.7}{0.6} & \result{64.6}{2.3} & \result{39.0}{0.2} & \result{18.9}{0.6} & \result{25.4}{0.7} & \result{70.6}{0.3} & \result{50.5}{2.0} & \result{66.0}{0.3} & \result{66.9}{0.6} & \result{50.4}{0.4} \\
 & WikiText & \result{67.2}{0.2} & \result{33.3}{0.5} & \result{64.0}{1.8} & \result{38.1}{0.1} & \result{18.9}{0.8} & \result{26.4}{0.7} & \result{70.1}{0.2} & \result{51.0}{0.6} & \result{65.6}{0.4} & \result{64.3}{0.5} & \result{49.9}{0.2} \\
\cdashlinelr{2-13}
 & Vocabulary & \result{65.4}{0.4} & \result{31.7}{0.5} & \result{56.0}{1.7} & \result{36.6}{0.2} & \result{13.0}{0.4} & \result{24.5}{0.6} & \result{69.7}{0.6} & \result{51.6}{1.8} & \result{62.2}{0.6} & \result{59.5}{0.6} & \result{47.0}{0.3} \\
 & Cosmopedia & \result{66.0}{0.9} & \result{31.5}{0.7} & \result{66.6}{2.1} & \result{38.2}{0.2} & \result{16.6}{0.7} & \result{23.5}{0.5} & \result{69.5}{0.4} & \result{53.9}{2.1} & \result{64.3}{0.5} & \result{64.4}{0.4} & \result{49.4}{0.4} \\
 & Self-calibration & \result{67.6}{0.3} & \result{35.1}{0.6} & \result{68.8}{2.0} & \result{39.5}{0.1} & \result{18.7}{0.5} & \result{25.8}{0.4} & \result{70.1}{0.3} & \result{59.1}{3.5} & \result{65.7}{0.3} & \result{64.9}{1.3} & \result{51.5}{0.7} \\
\bottomrule
\end{tabular}
\caption{Task accuracy across five calibration sets for Phi-2 2.7B, with standard deviation denoted in subscript.}
\label{tab:results_tasks_phi_2}
\end{table*}

\begin{table*}[t]
\scriptsize
\centering
\begin{tabular}{llrrrrrrrrrrr}
\toprule
Method & Dataset & \multicolumn{1}{c}{ARC-e} & \multicolumn{1}{c}{ARC-c} & \multicolumn{1}{c}{BoolQ} & \multicolumn{1}{c}{HS} & \multicolumn{1}{c}{LMBD} & \multicolumn{1}{c}{OBQA} & \multicolumn{1}{c}{PIQA} & \multicolumn{1}{c}{RTE} & \multicolumn{1}{c}{SC} & \multicolumn{1}{c}{WG} & \multicolumn{1}{c}{Mean} \\
\midrule
- & - & \multicolumn{1}{c}{65.6} & \multicolumn{1}{c}{30.5} & \multicolumn{1}{c}{66.1} & \multicolumn{1}{c}{50.5} & \multicolumn{1}{c}{63.3} & \multicolumn{1}{c}{27.6} & \multicolumn{1}{c}{76.3} & \multicolumn{1}{c}{55.2} & \multicolumn{1}{c}{73.6} & \multicolumn{1}{c}{65.2} & \multicolumn{1}{c}{57.4} \\
\cmidrule{1-13}
\multirow[c]{5}{*}{AWQ} & C4 & \result{65.6}{0.1} & \result{30.8}{0.2} & \result{65.7}{0.5} & \result{50.1}{0.0} & \result{63.5}{0.1} & \result{27.4}{0.2} & \result{76.8}{0.2} & \result{56.7}{0.9} & \result{74.0}{0.3} & \result{65.0}{0.3} & \result{57.6}{0.1} \\
 & WikiText & \result{65.5}{0.2} & \result{30.8}{0.4} & \result{65.9}{0.5} & \result{50.1}{0.0} & \result{63.8}{0.2} & \result{27.1}{0.4} & \result{76.4}{0.1} & \result{56.0}{0.9} & \result{74.1}{0.2} & \result{65.1}{0.5} & \result{57.5}{0.1} \\
\cdashlinelr{2-13}
 & Vocabulary & \result{64.9}{0.4} & \result{31.0}{0.3} & \result{62.5}{2.2} & \result{49.8}{0.1} & \result{59.3}{1.8} & \result{27.6}{0.4} & \result{76.2}{0.3} & \result{57.0}{1.4} & \result{73.4}{0.2} & \result{64.3}{0.5} & \result{56.6}{0.3} \\
 & Cosmopedia & \result{65.4}{0.1} & \result{31.0}{0.2} & \result{66.0}{0.5} & \result{50.0}{0.1} & \result{64.0}{0.3} & \result{27.5}{0.4} & \result{76.8}{0.3} & \result{56.3}{0.4} & \result{74.1}{0.2} & \result{65.1}{0.5} & \result{57.6}{0.1} \\
 & Self-calibration & \result{65.8}{0.2} & \result{30.9}{0.3} & \result{65.6}{0.4} & \result{50.2}{0.1} & \result{63.6}{0.2} & \result{26.9}{0.5} & \result{77.1}{0.2} & \result{56.8}{0.7} & \result{73.9}{0.2} & \result{64.9}{0.4} & \result{57.6}{0.1} \\
\cmidrule{1-13}
\multirow[c]{5}{*}{GPTQ} & C4 & \result{64.9}{0.2} & \result{30.4}{0.3} & \result{65.3}{1.0} & \result{49.6}{0.1} & \result{62.8}{0.3} & \result{26.5}{0.5} & \result{75.9}{0.1} & \result{54.2}{1.1} & \result{73.2}{0.2} & \result{64.7}{0.5} & \result{56.8}{0.2} \\
 & WikiText & \result{64.7}{0.3} & \result{30.6}{0.2} & \result{65.5}{0.4} & \result{49.7}{0.1} & \result{62.8}{0.2} & \result{26.9}{0.4} & \result{76.1}{0.3} & \result{55.1}{0.7} & \result{73.2}{0.3} & \result{64.6}{0.4} & \result{56.9}{0.1} \\
\cdashlinelr{2-13}
 & Vocabulary & \result{65.0}{0.5} & \result{30.7}{0.4} & \result{64.4}{2.0} & \result{49.8}{0.1} & \result{60.4}{1.8} & \result{27.2}{0.4} & \result{76.1}{0.3} & \result{55.5}{1.2} & \result{72.7}{0.3} & \result{64.1}{0.6} & \result{56.6}{0.3} \\
 & Cosmopedia & \result{65.1}{0.3} & \result{30.2}{0.6} & \result{64.8}{0.7} & \result{49.6}{0.1} & \result{62.5}{0.4} & \result{27.2}{0.4} & \result{75.5}{0.3} & \result{55.3}{0.8} & \result{73.3}{0.3} & \result{64.5}{0.4} & \result{56.8}{0.1} \\
 & Self-calibration & \result{65.5}{0.2} & \result{30.3}{0.7} & \result{65.1}{0.4} & \result{49.8}{0.0} & \result{62.3}{0.5} & \result{26.9}{0.4} & \result{76.0}{0.3} & \result{55.2}{1.2} & \result{73.2}{0.2} & \result{64.7}{0.5} & \result{56.9}{0.2} \\
\cmidrule{1-13}
\multirow[c]{5}{*}{SparseGPT} & C4 & \result{59.6}{0.3} & \result{25.4}{0.7} & \result{63.0}{0.4} & \result{43.2}{0.1} & \result{55.2}{0.8} & \result{23.9}{0.5} & \result{72.4}{0.6} & \result{53.1}{0.4} & \result{70.0}{0.3} & \result{61.8}{0.5} & \result{52.8}{0.2} \\
 & WikiText & \result{59.1}{0.9} & \result{26.2}{0.5} & \result{62.1}{0.1} & \result{41.3}{0.2} & \result{50.6}{0.5} & \result{24.4}{0.4} & \result{70.1}{0.6} & \result{52.9}{0.5} & \result{68.1}{0.2} & \result{61.3}{1.5} & \result{51.6}{0.2} \\
\cmidrule(lr){2-13}
 & Vocabulary & \result{54.4}{0.5} & \result{22.8}{0.4} & \result{62.4}{0.3} & \result{38.4}{0.1} & \result{38.1}{1.2} & \result{17.7}{0.5} & \result{70.3}{0.5} & \result{52.6}{0.5} & \result{63.9}{0.5} & \result{56.3}{1.1} & \result{47.7}{0.2} \\
 & Cosmopedia & \result{59.9}{0.6} & \result{26.3}{0.6} & \result{62.2}{0.0} & \result{42.3}{0.3} & \result{41.6}{0.7} & \result{24.8}{0.6} & \result{71.9}{0.4} & \result{53.1}{0.4} & \result{67.4}{0.7} & \result{60.0}{0.8} & \result{50.9}{0.2} \\
 & Self-calibration & \result{58.6}{0.4} & \result{25.8}{0.8} & \result{65.3}{0.9} & \result{42.2}{0.2} & \result{55.6}{0.5} & \result{23.9}{0.4} & \result{71.9}{0.6} & \result{52.6}{1.1} & \result{69.7}{0.4} & \result{60.9}{0.6} & \result{52.7}{0.3} \\
\cmidrule{1-13}
\multirow[c]{5}{*}{Wanda} & C4 & \result{56.7}{0.5} & \result{24.7}{0.4} & \result{62.3}{0.1} & \result{41.6}{0.1} & \result{43.9}{0.2} & \result{23.2}{0.9} & \result{71.2}{0.3} & \result{53.7}{0.3} & \result{68.4}{0.4} & \result{60.2}{0.7} & \result{50.6}{0.2} \\
 & WikiText & \result{56.0}{0.1} & \result{24.8}{0.4} & \result{62.2}{0.0} & \result{39.6}{0.2} & \result{40.2}{0.4} & \result{21.5}{0.8} & \result{69.8}{0.4} & \result{53.1}{0.3} & \result{66.4}{0.4} & \result{58.7}{0.4} & \result{49.2}{0.2} \\
\cdashlinelr{2-13}
 & Vocabulary & \result{47.4}{0.2} & \result{20.4}{0.4} & \result{62.2}{0.0} & \result{33.4}{0.1} & \result{22.4}{0.4} & \result{14.4}{0.1} & \result{66.6}{0.1} & \result{53.9}{1.4} & \result{58.7}{0.3} & \result{52.8}{0.7} & \result{43.2}{0.1} \\
 & Cosmopedia & \result{57.0}{0.1} & \result{24.7}{0.3} & \result{62.2}{0.0} & \result{40.7}{0.2} & \result{31.0}{0.7} & \result{23.0}{0.5} & \result{70.9}{0.4} & \result{52.7}{0.0} & \result{66.0}{0.5} & \result{58.5}{0.4} & \result{48.7}{0.2} \\
 & Self-calibration & \result{56.3}{0.2} & \result{24.6}{0.3} & \result{64.1}{0.4} & \result{41.3}{0.1} & \result{45.7}{0.5} & \result{21.5}{0.3} & \result{70.8}{0.4} & \result{53.9}{0.3} & \result{68.3}{0.3} & \result{60.1}{0.7} & \result{50.7}{0.2} \\
\bottomrule
\end{tabular}
\caption{Task accuracy across five calibration sets for OPT 6.7B, with standard deviation denoted in subscript.}
\label{tab:results_tasks_opt_6.7b}
\end{table*}

\begin{table*}[t]
\scriptsize
\centering
\begin{tabular}{llrrrrrrrrrrr}
\toprule
Method & Dataset & \multicolumn{1}{c}{ARC-e} & \multicolumn{1}{c}{ARC-c} & \multicolumn{1}{c}{BoolQ} & \multicolumn{1}{c}{HS} & \multicolumn{1}{c}{LMBD} & \multicolumn{1}{c}{OBQA} & \multicolumn{1}{c}{PIQA} & \multicolumn{1}{c}{RTE} & \multicolumn{1}{c}{SC} & \multicolumn{1}{c}{WG} & \multicolumn{1}{c}{Mean} \\
\midrule
- & - & \multicolumn{1}{c}{79.6} & \multicolumn{1}{c}{48.7} & \multicolumn{1}{c}{82.4} & \multicolumn{1}{c}{60.9} & \multicolumn{1}{c}{69.2} & \multicolumn{1}{c}{33.6} & \multicolumn{1}{c}{80.3} & \multicolumn{1}{c}{67.9} & \multicolumn{1}{c}{78.3} & \multicolumn{1}{c}{73.6} & \multicolumn{1}{c}{67.4} \\
\cmidrule{1-13}
\multirow[c]{5}{*}{AWQ} & C4 & \result{79.3}{0.0} & \result{48.1}{0.2} & \result{81.6}{0.4} & \result{59.9}{0.1} & \result{68.1}{0.4} & \result{33.6}{0.6} & \result{79.7}{0.2} & \result{69.2}{0.4} & \result{78.3}{0.1} & \result{72.9}{0.2} & \result{67.1}{0.0} \\
 & WikiText & \result{79.4}{0.3} & \result{48.0}{0.6} & \result{81.8}{0.4} & \result{60.0}{0.1} & \result{68.1}{0.2} & \result{33.7}{0.4} & \result{79.9}{0.2} & \result{69.1}{0.8} & \result{78.5}{0.3} & \result{72.4}{0.5} & \result{67.1}{0.1} \\
\cdashlinelr{2-13}
 & Vocabulary & \result{79.3}{0.3} & \result{48.3}{0.2} & \result{79.3}{0.5} & \result{59.9}{0.1} & \result{67.1}{0.6} & \result{33.4}{0.5} & \result{79.7}{0.1} & \result{67.4}{0.5} & \result{77.8}{0.4} & \result{72.3}{0.7} & \result{66.5}{0.1} \\
 & Cosmopedia & \result{79.3}{0.3} & \result{48.6}{0.1} & \result{81.6}{0.4} & \result{60.0}{0.1} & \result{67.6}{0.2} & \result{34.0}{0.5} & \result{79.4}{0.2} & \result{67.5}{1.2} & \result{78.9}{0.2} & \result{72.8}{0.4} & \result{67.0}{0.2} \\
 & Self-calibration & \result{79.1}{0.2} & \result{47.9}{0.9} & \result{81.8}{0.3} & \result{60.0}{0.1} & \result{68.0}{0.2} & \result{34.2}{0.5} & \result{80.1}{0.2} & \result{68.2}{1.2} & \result{78.5}{0.2} & \result{72.7}{0.3} & \result{67.0}{0.2} \\
\cmidrule{1-13}
\multirow[c]{5}{*}{GPTQ} & C4 & \result{79.0}{0.3} & \result{48.0}{0.5} & \result{81.8}{0.7} & \result{60.0}{0.2} & \result{68.0}{0.6} & \result{32.7}{0.2} & \result{80.1}{0.3} & \result{67.1}{2.1} & \result{78.3}{0.3} & \result{73.1}{0.5} & \result{66.8}{0.3} \\
 & WikiText & \result{79.1}{0.4} & \result{47.9}{0.5} & \result{82.1}{0.5} & \result{60.1}{0.1} & \result{68.0}{0.5} & \result{32.2}{0.7} & \result{80.0}{0.3} & \result{67.5}{1.7} & \result{78.4}{0.4} & \result{73.4}{0.4} & \result{66.9}{0.3} \\
\cdashlinelr{2-13}
 & Vocabulary & \result{78.4}{0.4} & \result{47.1}{1.0} & \result{81.6}{0.3} & \result{59.9}{0.1} & \result{67.0}{0.6} & \result{32.5}{0.6} & \result{79.7}{0.2} & \result{63.9}{1.7} & \result{77.2}{0.2} & \result{72.5}{0.5} & \result{66.0}{0.1} \\
 & Cosmopedia & \result{79.1}{0.3} & \result{47.1}{0.5} & \result{81.5}{0.6} & \result{60.0}{0.1} & \result{67.9}{0.2} & \result{32.0}{0.4} & \result{80.2}{0.3} & \result{66.7}{2.1} & \result{78.1}{0.4} & \result{73.1}{0.4} & \result{66.6}{0.2} \\
 & Self-calibration & \result{78.3}{0.3} & \result{46.8}{0.5} & \result{80.7}{0.9} & \result{59.9}{0.2} & \result{66.9}{0.6} & \result{32.0}{0.6} & \result{79.4}{0.4} & \result{63.0}{0.7} & \result{78.3}{0.2} & \result{73.1}{0.4} & \result{65.9}{0.2} \\
\cmidrule{1-13}
\multirow[c]{5}{*}{SparseGPT} & C4 & \result{67.4}{0.7} & \result{34.3}{0.8} & \result{75.2}{0.9} & \result{46.7}{0.3} & \result{53.9}{0.6} & \result{23.9}{0.5} & \result{73.3}{0.7} & \result{60.3}{1.9} & \result{71.8}{0.5} & \result{66.3}{0.9} & \result{57.3}{0.3} \\
 & WikiText & \result{67.2}{0.4} & \result{33.3}{0.5} & \result{64.3}{0.2} & \result{45.2}{0.2} & \result{54.0}{0.5} & \result{23.1}{0.4} & \result{71.3}{0.3} & \result{60.1}{2.5} & \result{70.4}{0.4} & \result{66.3}{0.6} & \result{55.5}{0.3} \\
\cdashlinelr{2-13}
 & Vocabulary & \result{62.5}{1.4} & \result{30.0}{0.8} & \result{71.0}{0.6} & \result{44.5}{0.2} & \result{42.7}{0.4} & \result{20.2}{0.6} & \result{71.5}{0.3} & \result{56.9}{2.9} & \result{68.6}{0.3} & \result{62.2}{0.9} & \result{53.0}{0.4} \\
 & Cosmopedia & \result{69.5}{0.4} & \result{35.4}{0.7} & \result{66.4}{0.8} & \result{45.6}{0.3} & \result{42.9}{0.8} & \result{24.2}{0.7} & \result{71.7}{0.2} & \result{60.7}{3.7} & \result{69.5}{0.5} & \result{64.9}{0.5} & \result{55.1}{0.3} \\
 & Self-calibration & \result{65.9}{0.5} & \result{32.2}{0.8} & \result{76.0}{0.9} & \result{46.7}{0.1} & \result{51.7}{0.8} & \result{23.2}{0.5} & \result{73.1}{0.6} & \result{60.5}{1.0} & \result{72.5}{0.2} & \result{66.5}{0.4} & \result{56.8}{0.3} \\
\cmidrule{1-13}
\multirow[c]{5}{*}{Wanda} & C4 & \result{64.3}{0.4} & \result{30.5}{0.6} & \result{70.4}{0.6} & \result{44.3}{0.1} & \result{42.3}{0.3} & \result{21.2}{0.6} & \result{71.9}{0.3} & \result{56.4}{2.0} & \result{70.8}{0.3} & \result{64.5}{0.5} & \result{53.7}{0.3} \\
 & WikiText & \result{64.8}{0.6} & \result{31.0}{0.5} & \result{66.1}{0.8} & \result{43.5}{0.1} & \result{44.5}{0.2} & \result{21.6}{0.4} & \result{70.7}{0.2} & \result{58.6}{1.1} & \result{70.0}{0.2} & \result{63.6}{0.5} & \result{53.4}{0.2} \\
\cdashlinelr{2-13}
 & Vocabulary & \result{58.4}{0.5} & \result{26.1}{0.2} & \result{64.3}{0.7} & \result{39.5}{0.2} & \result{29.8}{0.3} & \result{17.8}{0.6} & \result{69.7}{0.1} & \result{54.6}{1.4} & \result{64.4}{0.2} & \result{59.0}{0.7} & \result{48.4}{0.2} \\
 & Cosmopedia & \result{65.5}{0.2} & \result{32.4}{0.2} & \result{65.1}{0.6} & \result{43.6}{0.1} & \result{37.6}{0.3} & \result{21.0}{0.7} & \result{70.7}{0.3} & \result{58.1}{1.6} & \result{69.3}{0.2} & \result{63.4}{0.5} & \result{52.7}{0.2} \\
 & Self-calibration & \result{63.7}{0.3} & \result{30.0}{0.6} & \result{68.6}{1.4} & \result{44.3}{0.1} & \result{41.7}{0.3} & \result{20.6}{0.7} & \result{71.6}{0.4} & \result{58.9}{0.6} & \result{70.8}{0.3} & \result{64.8}{0.4} & \result{53.5}{0.1} \\
\bottomrule
\end{tabular}
\caption{Task accuracy across five calibration sets for Mistral 7B, with standard deviation denoted in subscript.}
\label{tab:results_tasks_mistral_7b_v0.3}
\end{table*}

\begin{table*}[t]
\scriptsize
\centering
\begin{tabular}{llrrrrrrrrrrr}
\toprule
Method & Dataset & \multicolumn{1}{c}{ARC-e} & \multicolumn{1}{c}{ARC-c} & \multicolumn{1}{c}{BoolQ} & \multicolumn{1}{c}{HS} & \multicolumn{1}{c}{LMBD} & \multicolumn{1}{c}{OBQA} & \multicolumn{1}{c}{PIQA} & \multicolumn{1}{c}{RTE} & \multicolumn{1}{c}{SC} & \multicolumn{1}{c}{WG} & \multicolumn{1}{c}{Mean} \\
\midrule
- & - & \multicolumn{1}{c}{81.4} & \multicolumn{1}{c}{51.5} & \multicolumn{1}{c}{82.2} & \multicolumn{1}{c}{60.0} & \multicolumn{1}{c}{67.1} & \multicolumn{1}{c}{33.4} & \multicolumn{1}{c}{80.0} & \multicolumn{1}{c}{70.0} & \multicolumn{1}{c}{78.2} & \multicolumn{1}{c}{74.0} & \multicolumn{1}{c}{67.8} \\
\cmidrule{1-13}
\multirow[c]{5}{*}{AWQ} & C4 & \result{80.7}{0.6} & \result{50.3}{0.6} & \result{79.9}{0.6} & \result{58.6}{0.1} & \result{66.1}{0.8} & \result{34.6}{0.4} & \result{79.5}{0.4} & \result{68.3}{0.9} & \result{77.3}{0.3} & \result{73.4}{0.4} & \result{66.9}{0.2} \\
 & WikiText & \result{80.4}{0.3} & \result{50.4}{0.9} & \result{81.3}{0.8} & \result{58.8}{0.1} & \result{65.6}{0.2} & \result{34.5}{0.3} & \result{79.5}{0.2} & \result{68.6}{0.5} & \result{77.7}{0.5} & \result{73.8}{0.3} & \result{67.1}{0.1} \\
\cdashlinelr{2-13}
 & Vocabulary & \result{80.3}{0.4} & \result{49.0}{1.0} & \result{80.0}{0.5} & \result{58.2}{0.2} & \result{63.4}{0.4} & \result{34.1}{0.8} & \result{79.2}{0.2} & \result{66.3}{0.9} & \result{76.6}{0.2} & \result{72.3}{0.2} & \result{66.0}{0.3} \\
 & Cosmopedia & \result{81.1}{0.1} & \result{50.3}{0.7} & \result{81.0}{0.5} & \result{58.8}{0.1} & \result{65.4}{0.5} & \result{34.7}{0.5} & \result{79.6}{0.2} & \result{67.1}{2.3} & \result{77.6}{0.4} & \result{73.0}{0.5} & \result{66.9}{0.3} \\
 & Self-calibration & \result{80.8}{0.3} & \result{50.0}{0.6} & \result{80.6}{0.2} & \result{58.7}{0.1} & \result{65.1}{0.4} & \result{34.5}{0.6} & \result{79.6}{0.3} & \result{66.1}{2.1} & \result{77.3}{0.3} & \result{73.7}{0.3} & \result{66.6}{0.3} \\
\cmidrule{1-13}
\multirow[c]{5}{*}{GPTQ} & C4 & \result{80.5}{0.4} & \result{48.7}{0.8} & \result{80.5}{0.4} & \result{58.8}{0.3} & \result{65.4}{0.7} & \result{32.7}{1.2} & \result{79.6}{0.3} & \result{71.9}{1.4} & \result{78.1}{0.3} & \result{72.8}{1.0} & \result{66.9}{0.3} \\
 & WikiText & \result{80.4}{0.4} & \result{48.5}{1.0} & \result{80.4}{0.5} & \result{58.8}{0.1} & \result{65.3}{0.2} & \result{33.2}{0.8} & \result{79.3}{0.1} & \result{69.5}{1.6} & \result{77.7}{0.4} & \result{72.9}{0.2} & \result{66.6}{0.3} \\
\cdashlinelr{2-13}
 & Vocabulary & \result{80.0}{0.5} & \result{47.9}{1.2} & \result{80.9}{0.6} & \result{58.0}{0.2} & \result{62.8}{0.5} & \result{33.8}{0.4} & \result{79.2}{0.5} & \result{65.1}{1.0} & \result{76.7}{0.3} & \result{72.6}{0.5} & \result{65.7}{0.1} \\
 & Cosmopedia & \result{80.8}{0.7} & \result{49.1}{0.4} & \result{81.1}{0.7} & \result{58.9}{0.1} & \result{64.7}{0.7} & \result{34.0}{0.9} & \result{79.2}{0.3} & \result{70.3}{1.0} & \result{77.9}{0.4} & \result{73.4}{0.4} & \result{66.9}{0.1} \\
 & Self-calibration & \result{79.7}{0.5} & \result{47.4}{0.8} & \result{81.4}{0.3} & \result{58.5}{0.3} & \result{64.9}{0.3} & \result{30.8}{0.8} & \result{79.2}{0.3} & \result{69.5}{1.7} & \result{77.7}{0.3} & \result{72.3}{0.4} & \result{66.1}{0.3} \\
\cmidrule{1-13}
\multirow[c]{5}{*}{SparseGPT} & C4 & \result{63.4}{0.9} & \result{31.0}{1.1} & \result{71.3}{1.8} & \result{43.9}{0.4} & \result{50.9}{0.6} & \result{23.0}{1.0} & \result{70.7}{0.6} & \result{57.4}{1.5} & \result{70.4}{0.4} & \result{65.9}{0.5} & \result{54.8}{0.3} \\
 & WikiText & \result{62.1}{1.0} & \result{30.0}{1.3} & \result{66.6}{2.2} & \result{41.4}{0.1} & \result{49.1}{1.4} & \result{22.6}{1.3} & \result{68.3}{0.4} & \result{53.5}{0.8} & \result{68.7}{0.5} & \result{63.9}{1.3} & \result{52.6}{0.4} \\
\cdashlinelr{2-13}
 & Vocabulary & \result{57.0}{0.5} & \result{25.6}{1.1} & \result{65.3}{1.4} & \result{37.2}{0.2} & \result{28.9}{1.2} & \result{20.1}{0.7} & \result{69.2}{0.5} & \result{52.6}{0.3} & \result{61.3}{0.5} & \result{56.1}{0.8} & \result{47.3}{0.4} \\
 & Cosmopedia & \result{64.6}{1.0} & \result{31.9}{1.1} & \result{64.8}{1.1} & \result{41.6}{0.4} & \result{34.3}{0.9} & \result{21.8}{1.3} & \result{68.9}{0.5} & \result{53.6}{0.5} & \result{66.4}{0.7} & \result{61.6}{1.1} & \result{50.9}{0.3} \\
 & Self-calibration & \result{64.5}{0.8} & \result{32.3}{1.0} & \result{70.7}{0.9} & \result{42.8}{0.2} & \result{43.3}{1.5} & \result{22.0}{0.9} & \result{69.5}{0.3} & \result{58.9}{2.4} & \result{69.6}{0.5} & \result{64.1}{0.5} & \result{53.8}{0.4} \\
\cmidrule{1-13}
\multirow[c]{5}{*}{Wanda} & C4 & \result{57.7}{0.6} & \result{26.8}{0.3} & \result{66.9}{1.6} & \result{38.2}{0.1} & \result{33.9}{0.5} & \result{19.3}{0.7} & \result{68.6}{0.2} & \result{53.5}{0.9} & \result{65.6}{0.2} & \result{59.8}{0.4} & \result{49.0}{0.3} \\
 & WikiText & \result{57.9}{0.4} & \result{28.2}{0.5} & \result{66.9}{0.3} & \result{37.8}{0.2} & \result{34.6}{0.4} & \result{20.4}{0.3} & \result{67.8}{0.3} & \result{53.1}{0.3} & \result{65.2}{0.3} & \result{59.8}{0.5} & \result{49.2}{0.1} \\
\cdashlinelr{2-13}
 & Vocabulary & \result{53.6}{0.4} & \result{22.9}{0.6} & \result{62.3}{0.2} & \result{34.0}{0.1} & \result{20.0}{1.2} & \result{18.1}{0.9} & \result{66.1}{0.5} & \result{52.1}{1.3} & \result{60.4}{0.4} & \result{57.3}{1.1} & \result{44.7}{0.3} \\
 & Cosmopedia & \result{57.7}{0.5} & \result{26.1}{0.2} & \result{65.7}{0.6} & \result{37.2}{0.2} & \result{26.6}{0.4} & \result{20.0}{0.8} & \result{68.6}{0.4} & \result{52.4}{0.7} & \result{64.0}{0.3} & \result{58.7}{0.6} & \result{47.7}{0.2} \\
 & Self-calibration & \result{58.1}{0.4} & \result{27.5}{0.2} & \result{67.7}{0.4} & \result{37.8}{0.2} & \result{31.8}{0.4} & \result{19.9}{0.7} & \result{69.0}{0.4} & \result{54.3}{1.3} & \result{65.7}{0.4} & \result{59.2}{0.5} & \result{49.1}{0.1} \\
\bottomrule
\end{tabular}
\caption{Task accuracy across five calibration sets for Llama 3.1 8B, with standard deviation denoted in subscript.}
\label{tab:results_tasks_llama_3.1_8b}
\end{table*}

\end{document}